\documentclass[11pt, a4paper, copyright,noabstract, nonumbering]{document}
\usepackage[authoryear, sort&compress, round]{natbib}
\usepackage{dblfloatfix}
\usepackage{ulem}
\usepackage{caption}
\usepackage{dramatist}
\usepackage{xspace}
\usepackage{pifont} %
\usepackage{multirow}
\usepackage{tcolorbox}
\usepackage{xltabular}
\usepackage{longtable}
\usepackage[hang, flushmargin]{footmisc}
\usepackage{float}
\hypersetup{
 colorlinks=true,
 linkcolor=blue,
 citecolor=blue,
 filecolor=blue,
 urlcolor=blue,
 pdfpagemode=UseNone,
 pdfstartview=FitH
}
\usepackage{threeparttable} 
\usepackage{fancyhdr}

\fancypagestyle{firstpage}{
    \fancyhf{} % 清空当前页面的所有页眉页脚
    \fancyhead[L]{\textbf{AgentSkiller Technical Report}} % 仅在左上角显示文字
    \fancyfoot[C]{\thepage} % 页脚居中显示页码
}

\usepackage{amsfonts}
\usepackage{amsmath}
\usepackage{amssymb}
\usepackage{lineno}
\usepackage{multirow}
\usepackage{adjustbox}

\usepackage{CJKutf8}
\usepackage{setspace}

\usepackage{dsfont}
\usepackage{array} %
\usepackage{tabularx} %
\usepackage{xcolor} %
\usepackage{tabularx}
\usepackage{booktabs}

\usepackage{lipsum}  %
\usepackage{multicol} %
\usepackage{subcaption}
\usepackage{listings}
\usepackage{xcolor}
\usepackage{tcolorbox}
\usepackage{caption}
\tcbuselibrary{listings, skins, breakable}

\usepackage{tablefootnote}
\usepackage{hyperref}

\lstdefinelanguage{json}{
 basicstyle=\ttfamily\footnotesize,
 numbers=left,
 numberstyle=\tiny\color{gray},
 stepnumber=1,
 numbersep=8pt,
 showstringspaces=false,
 breaklines=true,
 frame=lines,
 backgroundcolor=\color{gray!10},
 morestring=[b]",
 literate=
  *{0}{{{\color{black}0}}}{1}
{1}{{{\color{black}1}}}{1}
{2}{{{\color{black}2}}}{1}
{3}{{{\color{black}3}}}{1}
{4}{{{\color{black}4}}}{1}
{5}{{{\color{black}5}}}{1}
{6}{{{\color{black}6}}}{1}
{7}{{{\color{black}7}}}{1}
{8}{{{\color{black}8}}}{1}
{9}{{{\color{black}9}}}{1}
}

\makeatletter
\def\@BTrule[#1]{%
  \ifx\longtable\undefined
 \let\@BTswitch\@BTnormal
  \else\ifx\hline\LT@hline
 \nobreak
 \let\@BTswitch\@BLTrule
  \else
  \let\@BTswitch\@BTnormal
  \fi\fi
  \global\@thisrulewidth=#1\relax
  \ifnum\@thisruleclass=\tw@\vskip\@aboverulesep\else
  \ifnum\@lastruleclass=\z@\vskip\@aboverulesep\else
  \ifnum\@lastruleclass=\@ne\vskip\doublerulesep\fi\fi\fi
  \@BTswitch}
\makeatother

\addto\extrasenglish{
}

 {\begin{list}{}%
 {\setlength{\leftmargin}{#1}}%
 \item[]%
 }
 {\end{list}}
 
\renewcommand{\today}{}

\title{\vspace{-0.2cm}\centering {AgentSkiller}: Scaling Generalist Agent Intelligence through Semantically Integrated Cross-Domain Data Synthesis}

\author[*]{
\small
Zexu Sun$^{\dagger,*,1}$, Bokai Ji$^{*,2}$, Hengyi Cai$^{1}$, Shuaiqiang Wang$^{1}$, Lei Wang$^{3}$, Guangxia Li$^{2}$, Xu Chen$^{4}$
\\
\small
{$^1$ AMU, Baidu Inc.}
{$^2$ School of Computer Science and Technology, Xidian University}
\\
\small
{$^3$ Singapore Management University}
{$^4$ Gaoling School of Artificial Intelligence, Renmin University of China}
\\
\vspace{0.1cm}
\small
$^{\dagger}$ Project Leader, \texttt{sunzexu0826@gmail.com}, $^{*}$ Equal Contribution  
\vspace{0.2cm}
  \\
  \small
  \includegraphics[height=0.9em]{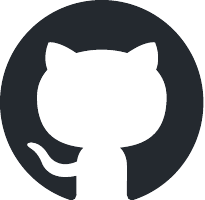} \textbf{Source Code}: \url{https://github.com/ZexuSun/AgentSkiller} \\
  \small
  \includegraphics[height=1.0em]{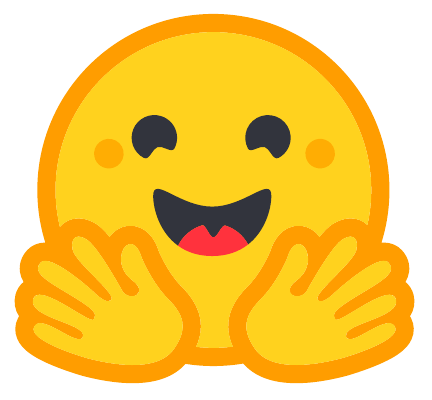} \textbf{Dataset \& Models}: \url{https://huggingface.co/AgentSkiller} \\
}

\renewcommand{\phi}{\varphi}

\renewcommand{\epsilon}{\varepsilon}
\renewcommand{\imath}{\mathrm{i}}

\newlength{\restsubwidth}
\newlength{\restsubheight}
\newlength{\restsubmoreheight}
\newcommand{\rest}[2]{%
        \settowidth{\restsubwidth}{\ensuremath{#2}}
        \settoheight{\restsubheight}{\ensuremath{{}_{#2}}}
        \ensuremath{{#1\hskip 0.5pt}_{\vrule\kern2pt\parbox[b][%
        4pt][b]{\the\restsubwidth}{%
                        \ensuremath{{}_{#2}}}}}
        }

\begin{document}
\maketitle
\thispagestyle{firstpage}
\vspace{-0.5cm}
\textbf{The rapid evolution of Large Language Model agents has demonstrated potential in solving complex real-world problems via external tools. However, advancing generalist agent intelligence is bottlenecked by the scarcity of high-quality, long-horizon training data. Existing synthesis methods fall into two categories: (1) collecting real-world API logs, often constrained by privacy and coverage, and (2) generating synthetic data through scripted interactions, which require manual validation and lack conversational diversity. Both approaches struggle to produce the data requisite for scaling generalist capabilities.}
%==
\textbf{To address these limitations, we propose \textbf{AgentSkiller}, a fully automated, scalable framework synthesizing multi-turn interaction data across realistic, {semantically linked} domains. AgentSkiller employs a reliable architecture based on Directed Acyclic Graph organized explicit state transitions to ensure determinism and recoverability.}%== The framework utilizes three core design principles: a \textit{Dual-Model Architecture} splitting planning from coding to balance cost and performance; \textit{State Transition Orchestration} saving progress at every step for reliability; and a \textit{Test-Driven Self-Correction} mechanism automatically detecting and fixing bugs in the generation procedure.}

\noindent \textbf{The synthesis pipeline follows five parts to turn ideas into fully functional environments. First, it builds a domain ontology and a \textit{Person-Centric Entity Graph} to structure data around realistic scenarios. Second, it defines clear tool interfaces by generating \textit{Service Blueprints} for Model Context Protocol servers. Third, it populates the environment with consistent databases and strict \textit{Domain Policy}, ensuring data logic never conflicts. Fourth, a cross-domain fusion mechanism links services to simulate complex, multi-turn tasks. Finally, the pipeline creates user tasks by verifying solution paths with a planner and filtering them through execution-based validation, while a \textit{Persona-based Simulator} generates natural language queries. An automated \textit{Rollout} mechanism executes these queries to produce high-fidelity interaction trajectories. By expanding these scenarios, AgentSkiller produces reliable environments with clear state changes, creating a solid foundation for training next-generation generalist agents.} %== \textbf{Additionally, experimental results show that models trained on AgentSkiller datasets achieve SOTA performance, particularly with larger models (e.g., Qwen-14B).}
\textbf{Additionally, to demonstrate the effectiveness of AgentSkiller, we synthesized approximately 11K interaction data, experimental results indicate that models trained on this dataset achieve significant improvements on function calling scenario over baselines, particularly in larger parameter regimes.}
\begin{figure*}[!h]
    \centering
    \includegraphics[width=0.99\linewidth]{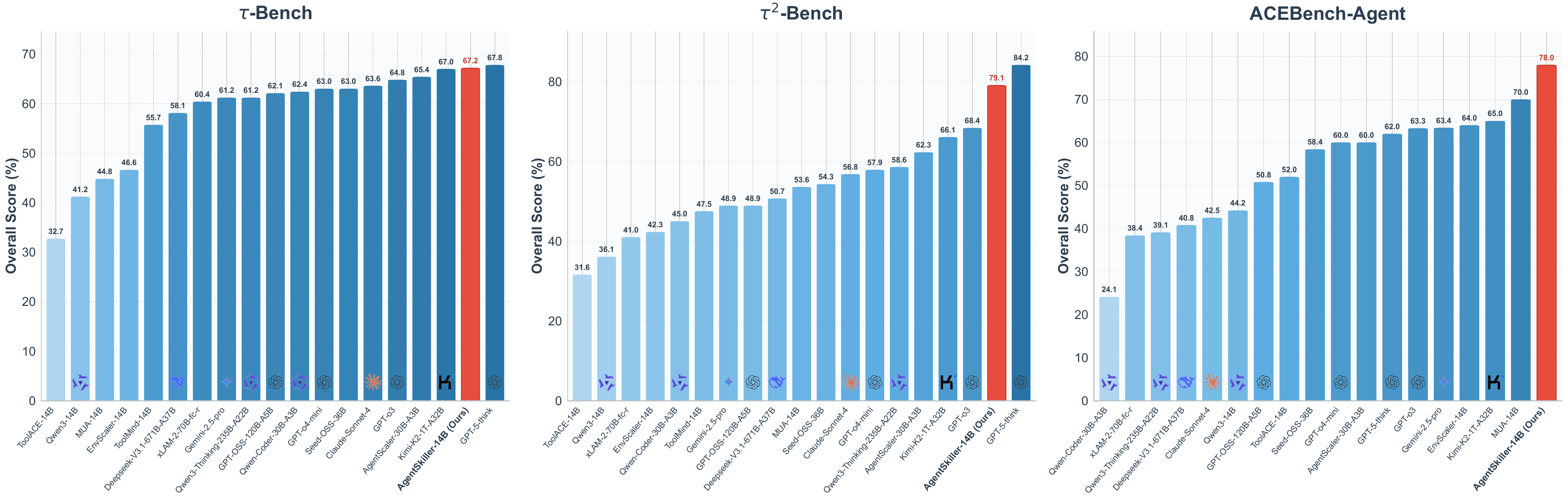}
    \captionsetup{justification=centering, singlelinecheck=false}
    \caption{Performance on $\tau$-bench, and $\tau^2$-bench and ACEBench-Agent. Specifically, our AgentSkiller-14B achieves significant improvement on $\tau^2$-bench and ACEBench-Agent.}
    \label{fig:placeholder}
\end{figure*}

\section{Introduction}
The rapid rise of tool-using agents is driven by the need for Large Language Models (LLMs) to handle complex tasks beyond simple writing~\citep{yao2022react,luo2025mcp, wang2025mcp}. 
By using function calling, LLMs can move past static text generation and operate across many specialized fields~\citep{schick2023toolformer,openai2023function}. 
To be truly useful, agents must accurately understand user intent, extract context, call external APIs, and manage multi-step conversations~\citep{patilberkeley}. 
These capabilities allow agents to access real-time data, creating a smooth user experience~\citep{kim2024llm,silver2025welcome,wu2025webdancer}. 
%==
However, acquiring reliable training data remains a critical bottleneck. This challenge is exacerbated in complex, cross-domain scenarios, where generating high-quality interaction trajectories that span diverse environments becomes increasingly difficult.

To address the issue of data scarcity, researchers typically employ two distinct strategies for generating synthetic agentic data.
%==
The first strategy adopts a reverse paradigm. In this approach, the system generates user queries to retrospectively match assistant function calls at each interaction turn. However, the resulting trajectories often fail to mimic realistic usage~\citep{yin2025magnet}.
%==
The second strategy follows a forward paradigm, also known as simulated agent–human interplay~\citep{chen2024facilitating, liu2024apigen, prabhakar2025apigen,barres2025tau}. Here, the process begins by formulating a high-level user intent to drive the interaction, effectively constructing agentic data in a ``top-down" manner. Despite this structured approach, the generated interactions frequently lack naturalness. 
%==
Furthermore, scalability remains a critical bottleneck. Without automated pipelines to construct these environments, large-scale deployment is impeded by the need for continuous manual intervention, rendering the generation of high-quality, multi-turn, cross-domain data infeasible.

More recently, efforts to automate environment construction have emerged. APIGen~\citep{liu2024apigen} collects 3,673 executable APIs across 21 categories to generate diverse function-calling datasets in a structured manner. 
AgentScaller~\citep{fang2025towards} designs a framework that automatically constructs heterogeneous, fully simulated environments to broaden function-calling scenarios. 
Similarly, ToolMind~\citep{yang2025toolmind} proposes a large-scale dataset with 160k synthetic instances using over 20k tools. 
%==
EnvScaler~\citep{song2026envscaler} builds automated framework for scalable tool-interaction environments via programmatic synthesis. 
Despite these advances, there is still a critical gap remains. Addressing generalist agent intelligence requires scalable training resources that expose agents to realistic, long-horizon, and diverse function-calling contexts. 
But, most of the existing synthesis approaches rely on collecting real-world API logs or generating simple synthetic interactions, are often constrained by limited coverage, privacy concerns, and a lack of high-quality multi-turn cross domain data. These limitations hinder the effective scaling of generalist agent intelligence.

To overcome these challenges, we propose \textbf{AgentSkiller}, a fully automated, scalable framework designed to synthesize multi-turn interaction data across realistic, high-fidelity domains. 
Unlike previous ad-hoc methods, AgentSkiller employs a robust state-machine-driven architecture orchestrated by a Directed Acyclic Graph (DAG) to ensure determinism and recoverability. 
Specifically, our AgentSkiller is built upon three core design principles: a \textit{Dual-Model Architecture} that decouples semantic reasoning from syntactic implementation, granular orchestration with automated checkpointing, and a \textit{Test-Driven Self-Correction} mechanism that iteratively corrects generated code. 
The pipeline systematically broadens the space of function-calling scenarios through five sequential phases: establishing domain ontologies with \textit{Person-Centric Entity Graphs}, standardizing Service Blueprints for Model Context Protocol (MCP) servers, implementing semantic grounding via constraint-satisfying databases, performing cross-domain fusion, and finally utilizing a \textit{Persona-Based Simulator} to generate natural language queries. 
By producing coherent, executable environments with deterministic state transitions, AgentSkiller provides a scalable foundation for synthesizing complex data to train next-generation generalist agents. 
%==
To rigorously validate the utility of the proposed framework, we synthesized a corpus comprising approximately 11k multi-turn interaction trajectories using AgentSkiller. Subsequent experiments across challenging function-calling benchmarks, including $\tau$-bench, $\tau^2$-bench and ACEBench, demonstrate that models trained on this dataset yield substantial performance gains. Notably, the AgentSkiller-14B exhibits exceptional capability in complex tool-use scenarios, consistently outperforming established open-source baselines and achieving parity with state-of-the-art proprietary models.

\section{Problem Formulation}

We formalize the multi-turn interaction between the assistant and the user as a Partially Observable Markov Decision Process (POMDP)~\citep{cassandra1998survey}, defined by the tuple $(\mathcal{U}, \mathcal{S}, \mathcal{A}, \mathcal{O}, \mathcal{P}, \mathcal{R}, \gamma)$.

\noindent
\textbf{User Intent Space ($\mathcal{U}$).}
Let $q \in \mathcal{U}$ denote the user's latent intent or goal. This intent acts as a hidden condition driving the user's responses. The assistant must infer $q$ incrementally through interaction history to generate appropriate actions.

\noindent
\textbf{State Space ($\mathcal{S}$).}
The global state space $\mathcal{S}$ is composed of two subspaces: the physical environment state $\mathcal{S}_{E}$ (e.g., database records, file systems) and the dialogue context state $\mathcal{S}_{H}$ (e.g., user's mental state and conversation history). At turn $t$, the full state $s_t = (s_{E}^t, s_{H}^t) \in \mathcal{S}$ encompasses all information about the environment and the interaction.

\noindent
\textbf{Action Space ($\mathcal{A}$).}
The assistant's action space $\mathcal{A} = \mathcal{A}_{\text{tool}} \cup \mathcal{A}_{\text{resp}}$ consists of two distinct types of actions:
\begin{itemize}
 \item \textbf{Tool Invocation ($a \in \mathcal{A}_{\text{tool}}$):} The assistant executes an external API call to manipulate the environment or retrieve information.
 \item \textbf{Response ($a \in \mathcal{A}_{\text{resp}}$):} The assistant generates a natural language response to communicate with the user.
\end{itemize}

\noindent
\textbf{Observation Space ($\mathcal{O}$).}
Since the true state $s_t$ is partially observable, the assistant receives an observation $o_t \in \mathcal{O}$ based on its action. The observation space is defined as $\mathcal{O} = \mathcal{O}_{E} \cup \mathcal{O}_{H}$, where $\mathcal{O}_{E}$ represents structured outputs from tool executions, and $\mathcal{O}_{H}$ represents natural language feedback from the user.

\noindent
\textbf{Transition Dynamics ($\mathcal{P}$).}
The transition function $\mathcal{P}: \mathcal{S} \times \mathcal{A} \to \Pi(\mathcal{S} \times \mathcal{O})$ defines the system dynamics. At turn $t$, given action $a_t$, the system transitions to $s_{t+1}$ and emits $o_{t+1}$:
\begin{itemize}
 \item If $a_t \in \mathcal{A}_{\text{tool}}$, the environment state updates $(s_{E}^t, a_t) \to s_{E}^{t+1}$, and the assistant observes a tool output $o_{E}^{t+1} \in \mathcal{O}_{E}$. The dialogue state $s_H$ remains effectively static regarding user intent, though the history is updated.
 \item If $a_t \in \mathcal{A}_{\text{resp}}$, the dialogue state updates $(s_{H}^t, a_t) \to s_{H}^{t+1}$, and the assistant observes a user reply $o_{H}^{t+1} \in \mathcal{O}_{H}$. The environment state $s_E$ remains static during this turn.
\end{itemize}
Crucially, the environment state $s_{E}$ remains latent; the agent can only infer it through observations $\mathcal{O}_{E}$.

\noindent
\textbf{Interaction Policy and Objective.}
The assistant operates according to a policy $\pi(a_t | h_t)$, where $h_t = (o_0, a_0, \dots, o_t)$ represents the interaction history. The interaction concludes at turn $T$ when either the user issues a termination signal or a maximum turn limit is reached.

The reward function $\mathcal{R}$ evaluates the quality of the trajectory $\tau = (s_0, a_0, \dots, s_T)$. We define the reward based on two criteria: (1) the effectiveness of the environment manipulation, measured by the cumulative state change $\Delta s_{E}$ relative to the user's intent $q$, and (2) the quality of the communicative interface. The assistant's objective is to find an optimal policy $\pi^*$ that maximizes the expected cumulative reward:
\begin{equation}
 \pi^* = \operatorname*{arg\,max}_{\pi} \mathbb{E}_{\tau \sim \pi} \left[ \sum_{t=0}^{T} \gamma^t \mathcal{R}(s_t, a_t, s_{t+1}) \right]
\end{equation}
where $\gamma \in [0, 1]$ is a discount factor. In our specific setting, the reward $\mathcal{R}$ is often sparse, calculated primarily upon the successful fulfillment of the user's intent $q$ (i.e., task success) and the helpfulness of the final response.

\section{AgentSkiller}

\subsection{Overview}
\label{sec:methodology_overview}

To operationalize the objective defined above, we present \textbf{AgentSkiller}, a robust and modular framework designed to automate the generation of high-fidelity, cross-domain, function-calling task environments for Large Language Model (LLM) agents. Furthermore, the AgentSkiller framework is capable of generating sophisticated cross-domain multi-turn dialogue datasets designed to simulate complex tasks with extensive interaction depth. Specifically, it adopts a state-machine-driven design orchestrated by a Directed Acyclic Graph (DAG), ensuring determinism and recoverability throughout the generation lifecycle.

\subsubsection{Architectural Design Principles}
The architecture of AgentSkiller is built upon three core design principles to ensure stability and quality in large-scale synthesis:

\begin{enumerate}
    \item \textbf{Dual-Model Architecture:} The framework decouples reasoning tasks from implementation tasks. A \textit{Textual LLM} (e.g., GPT-5.2) handles high-level semantic reasoning (e.g., domain expansion, policy formulation), while a specialized \textit{Coding LLM} (e.g., Claude Opus) is dedicated to syntactic precision (e.g., SQL generation, Python implementation). This separation optimizes both cost and performance.
    \item \textbf{State-Machine Orchestration:} The pipeline is structured as a sequence of 18 atomic steps managed by LangGraph\footnote{\url{https://github.com/langchain-ai/langgraph}}. Each step incorporates automated checkpointing, allowing the workflow to pause, resume, or recover from failures without data loss, which is a critical feature for long-running generation tasks.
    \item \textbf{Test-Driven Self-Correction:} {To mitigate the stochastic nature of generative models, a \textit{Test-Driven Self-Correction} mechanism is implemented. Syntax errors or logical violations in the generated code (e.g., in MCP servers or database scripts) trigger an automated ``Block Editor'' that iteratively patches the code based on error traces.}
\end{enumerate}

\subsubsection{Pipeline Phasing}
\begin{figure*}[!t] % 注意 Figure 2 很宽，建议用 figure* 跨栏显示，如果是单栏排版则用 figure
    \centering
    \includegraphics[width=\linewidth]{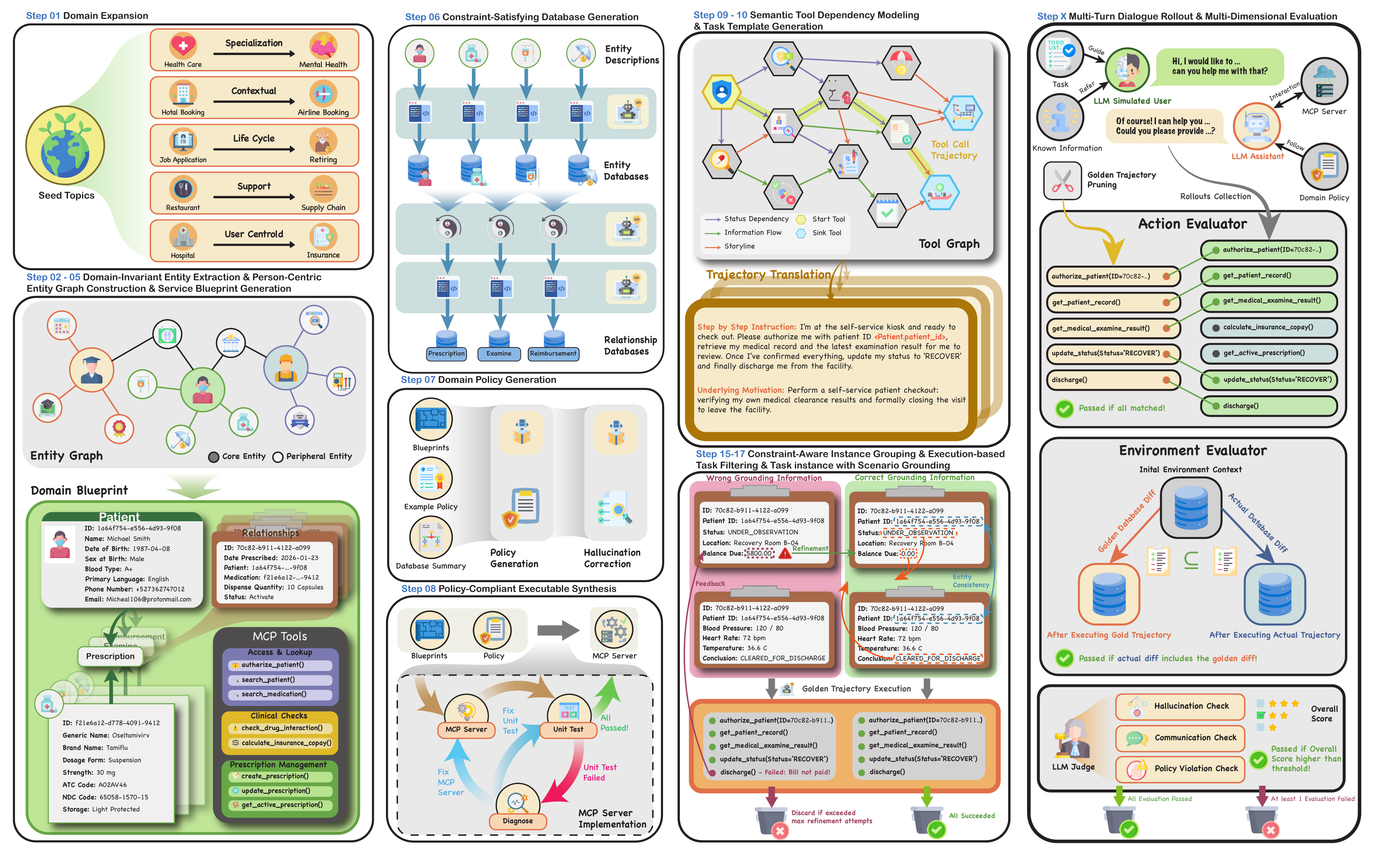} % 替换为你的文件名
   \caption{\textbf{The End-to-End Single-Domain Synthesis Framework.} The pipeline progresses through \textit{Ontology Definition} (Steps 1--5), \textit{Executable Implementation} (Steps 6--8), and \textit{Task Instantiation} (Steps 9--17). Crucially, the rightmost column details the \textit{Execution-Based Task Filtering} and \textit{Multi-Dimensional Evaluation}, ensuring only valid, high-fidelity environments are retained.}
    \label{fig:single_domain}
\end{figure*}

The AgentSkiller framework is logically organized into five sequential phases, transforming abstract seed concepts into fully instantiated, executable environments. As comprehensively illustrated in Figure~\ref{fig:single_domain}, this \textbf{End-to-End Single-Domain Synthesis Framework} operates as the foundational engine. The workflow follows a strict "Generate-Implement-Validate" progression:

\begin{itemize}
    \item \textbf{Ontology \& Blueprint (Left Panel):} The process begins with semantically expanding seed topics into a person-centric entity graph (Steps 1--5), defining the structural boundaries of the domain.
    \item \textbf{Executable Implementation (Center Panel):} Abstract definitions are then grounded into concrete databases and policy-compliant MCP servers (Steps 6--8).
    \item \textbf{Task Instantiation \& Validation (Right Panel):} Crucially, as depicted in the rightmost column of Figure~\ref{fig:single_domain}, the pipeline culminates in a rigorous \textit{Execution-Based Task Filtering} and \textit{Multi-Dimensional Evaluation} phase (Step 10 \& Steps 15--17). This ensures that only functionally robust, high-fidelity environments are retained as building blocks for the subsequent cross-domain fusion.
\end{itemize}

\noindent
\textbf{Phase 1: Domain Ontology and Topology (Steps 1--3)}
This phase establishes the semantic boundaries of the simulation. Starting from a curated set of seed domains (e.g., Healthcare, Finance), the system iteratively expands the ontology into a broad corpus of topics. It then extracts domain-invariant entities and constructs a \textit{Person-Centric Entity Graph}, modeling realistic relationships between human actors and system resources to ground the simulation in real-world interaction scenarios.

\noindent
\textbf{Phase 2: Service Specification (Steps 4--5)}
Abstract entities are formalized into concrete service definitions, termed \textit{Service Blueprints}, which correspond to Model Context Protocol (MCP) servers. This phase establishes the functional boundaries, listing available tools and data schemas. We enforce strict interface standardization, mapping abstract functions to executable function call schemas to ensure agent interoperability.

\noindent
\textbf{Phase 3: Implementation and Semantic Grounding (Steps 6--9)}
Transitioning from specification to execution, this phase synthesizes concrete databases populated with constraint-satisfying data, generates rigorous \textit{Domain Policy} documents to govern tool behavior, and implements fully functional Python MCP servers with accompanying unit tests. A key contribution here is the \textbf{Summary Constraint Generation}, which ensures relational integrity in the synthesized data.

\noindent
\textbf{Phase 4: Cross-Domain Fusion (Steps 10--13)}
Designed to simulate complex, real-world interoperability, this optional phase (executed after single-domain validation) links different services together to simulate complex, multi-turn tasks. The system fuses their respective databases and policies into a unified execution context, enabling multi-hop reasoning tasks across service boundaries.

\noindent
\textbf{Phase 5: Task Synthesis and Instantiation (Steps 14--17)}
The final phase generates the user-facing evaluation artifacts. It enumerates valid tool execution trajectories from the tool graph and grounds them in specific database instances via a constraint-solving planner. Subsequently, a \textit{Task Filtering} module executes these grounded trajectories in a sandboxed environment to eliminate runtime failures. Finally, it synthesizes natural language user queries using a \textit{Persona-Based Simulator}, ensuring the resulting tasks are semantically aligned with the simulated state and test for active information gathering.

This structured approach ensures that the resulting benchmarks are not merely collections of isolated prompts, but coherent, executable environments where agent actions have deterministic and stateful consequences.

\subsection{Domain Expansion (Step 1)}
As the inaugural step of AgentSkiller, the objective is to extrapolate a comprehensive set of domain topics from a limited set of high-level categories. This constructs a broad semantic foundation for the subsequent entity extraction and tool graph construction, serving as the cornerstone for generating cross-domain data.

\subsubsection{Problem Formulation}
Let $D_{\text{seed}}$ denote the initial set of seed domains. These seeds represent high-level industries such as \textit{healthcare}, \textit{finance}, and \textit{manufacturing}. The goal is to generate an expanded set $D_{\text{target}}$ such that $D_{\text{seed}} \subset D_{\text{target}}$, $D_{\text{seed}}=n$, and $D_{\text{target}} = N$, where $n$ and $N$ are pre-defined hyperparameters (in our implementation, $n=15$ and $N=300$). The expansion process is driven by the Textual LLM $\mathcal{M}$ (e.g., GPT-5.2), which functions as a semantic generator.

\subsubsection{Expansion Strategies}
To prevent the generated domains from converging into repetitive or trivial clusters, a prompting strategy is designed to guide $\mathcal{M}$ in exploring the semantic space across five distinct dimensions. This ensures that the resulting topology covers breadth, depth, and temporal relationships:

\begin{itemize}
    \item \textbf{Horizontal Association (Same-Context):} Identifying parallel sectors within the same broader industry (e.g., expanding \textit{hotel} to \textit{airline} and \textit{car\_rental}).
    \item \textbf{Diagonal Association (Cross-Context):} Bridging intersections between distinct industries (e.g., linking \textit{healthcare} with \textit{finance} to derive \textit{medical\_billing}).
    \item \textbf{Temporal Derivation (Lifecycle):} Mapping sequential dependencies within a workflow (e.g., \textit{job\_search} $\rightarrow$ \textit{employee\_onboarding}).
    \item \textbf{Hierarchical Refinement (Specialization):} Drilling down from general categories to specific sub-sectors (e.g., \textit{healthcare} $\rightarrow$ \textit{dental\_services}).
    \item \textbf{Agent-Centric Mapping (Role-Based):} Deriving domains based on specific operational roles (e.g., \textit{hospital} $\rightarrow$ \textit{nurse\_scheduling}).
\end{itemize}

\subsubsection{Implementation and Constraints}
The expansion process follows an iterative ``Generate-Verify-Store'' loop. At each iteration $t$, the model generates a batch of candidate domains $B_t$ (in our implementation, the batch size $k=20$).

Strict syntactic and semantic constraints are enforced to ensure the utility of the generated topics:
\begin{itemize}
    \item \textbf{Format:} All domain labels are standardized to \texttt{snake\_case} to facilitate downstream code generation.
    \item \textbf{Granularity:} Labels are restricted to 1--3 vocabulary terms to balance specificity against generalization, strictly avoiding overly broad terms (e.g., ``management'') or hyper-specific scenarios.
    \item \textbf{Deduplication:} A global set $S$ of unique domains is maintained. For every new batch $B_t$, the set is updated via $S \leftarrow S \cup B_t$, discarding any duplicates to maximize semantic entropy.
\end{itemize}

The process incorporates a checkpointing mechanism to support incremental generation and fault tolerance, ensuring the pipeline can resume from the last valid state until the target count $N$ is achieved.

\subsection{Domain-Invariant Entity Extraction (Step 2)}
Following the establishment of the domain corpus, the subsequent step involves distilling abstract domain topics into concrete, reusable data structures. This module, \textit{Entity Extraction}, aims to generate a standardized set of entities $\mathcal{E}$ and their associated attribute schemas from the expanded domain set $D_{\text{target}}$.

\subsubsection{Global Extraction Strategy}
Unlike traditional iterative extraction methods that process domains sequentially, we employ a \textbf{Global One-Shot Extraction} strategy. The entire set of domain topics $D_{\text{target}}$ is fed into the Textual LLM in a single context window. This approach allows the model to identify and consolidate overlapping concepts across different domains (e.g., recognizing that a \textit{Server} entity in the ``Cloud Computing'' domain shares fundamental properties with a \textit{Server} in ``IT Management''), thereby enforcing semantic consistency and reducing redundancy at the source.

\subsubsection{Domain-Invariant Granularity Constraint}
To ensure the generated entities are viable for complex relationship modeling, a theoretical constraint known as \textbf{Domain-Invariant Granularity} is imposed. This constraint dictates that a valid entity $e \in \mathcal{E}$ must satisfy three conditions:

\begin{itemize}
    \item \textbf{Domain Specificity vs. Universality:} The entity must be specific enough to hold meaningful attributes but broad enough to function across related sub-domains. For instance, generic terms like \textit{Person} are rejected for lack of specificity, while hyper-specific terms like \textit{MiddleSchoolStudent} are rejected for lack of generalization; \textit{Student} is accepted as the optimal mid-level granularity.
    \item \textbf{Existential Independence:} An entity must exist independently of other entities. Relational concepts (e.g., \textit{Enrollment}, which depends on \textit{Student} and \textit{Course}) are explicitly categorized as relationships rather than entities, and are deferred to the subsequent graph construction phase.
    \item \textbf{Semantic Stability:} The definition of the entity must remain consistent across distinct domains. Concepts that shift meaning based on context (e.g., \textit{TicketType}, which differs vastly between aviation and support desk contexts) are excluded.
\end{itemize}

\subsubsection{Attribute Schema}
Beyond basic identifiers, the Textual LLM is required to generate a high-dimensional feature set for each entity. To prevent the generation of superficial schemas, a mandatory expansion across four orthogonal dimensions is enforced: \textbf{Physical Specifications} (e.g., \textit{dimensions}, \textit{storage\_capacity}), \textbf{Lifecycle Temporal Data} (e.g., \textit{manufacturing\_year}, \textit{warranty\_expiration}), \textbf{State Configuration} (e.g., \textit{is\_active}, \textit{firmware\_version}, \textit{access\_level}), and \textbf{Geospatial/Network Addressing} (e.g., \textit{ip\_address}, \textit{geo\_coordinates}).

\subsubsection{Post-Sanitization Process}
The raw output from the Textual LLM undergoes a deterministic post-processing phase to ensure system compatibility. First, a \textbf{Namespace Collision Filter} is implemented to remove system-reserved entities (specifically \textit{Session}, which conflicts with the multi-turn dialogue architecture context). Second, a strict \textbf{Primary Key Normalization} rule is enforced, where the identifier for an entity $E$ is programmatically rewritten as $E_{\text{name}}.\text{lower}() + \text{``\_id''}$ (e.g., \texttt{LabEquipment} $\rightarrow$ \texttt{labequipment\_id}). This standardization is critical for the automated generation of foreign keys in the subsequent database synthesis step.

\subsection{Person-Centric Entity Graph Construction (Step 3)}
Building upon the extraction of discrete entities, the pipeline advances to the \textit{Entity Graph Generation} phase. The objective is to transition from a flat set of isolated definitions to a structured semantic topology. We construct a directed graph $G = (\mathcal{V}, \mathcal{L})$, where nodes $\mathcal{V}$ represent the extracted entities, and links $\mathcal{L}$ represent potential semantic interactions.

\subsubsection{Topology Definition}
To simulate realistic business scenarios, a \textbf{Person-Centric} architectural constraint is adopted. It is hypothesized that in enterprise simulations, operational blueprints are centered around human actors (e.g., \textit{Student}, \textit{Clinician}, \textit{Employee}) interacting with system resources (e.g., \textit{Course}, \textit{Appointment}, \textit{Project}).

Formally, let $\mathcal{E}$ be the set of all entities extracted in the previous step. We identify a subset of actors $\mathcal{P} \subset \mathcal{E}$ where the attribute \texttt{is\_person} is true. The edge generation process aims to identify a set of directed links $\mathcal{L} \subseteq \mathcal{P} \times (\mathcal{E} \setminus \{p\})$, such that an edge $(p, e)$ exists if and only if actor $p$ semantically operates upon or relates to entity $e$ within a specific domain context.

\subsubsection{LLM-Driven Relation Inference}
Determining semantic relevance between arbitrary entities requires common-sense reasoning that transcends static rule-based systems. We employ the Textual LLM as a zero-shot inference engine to judge the existence of relations.

The inference process is governed by a \textit{Domain Relevance Prompt}. For a given actor $p \in \mathcal{P}$ and a candidate target entity $e \in \mathcal{E}$, the model is queried to determine if $e$ is relevant to $p$ in any plausible domain scenario (e.g., judging that a \textit{Ticket} is relevant to a \textit{Passenger} in the Travel domain). This approach effectively maps the latent knowledge within the LLM into explicit graph edges, creating a ``hub-and-spoke'' local topology around each human actor.

\subsubsection{Scalable Execution Strategy}
Checking relations between every actor and every other entity implies a complexity of $O(|\mathcal{P}| \times |\mathcal{E}|)$. To handle the potential scale of generated entities, a robust execution engine is implemented featuring:

\begin{itemize}
    \item \textbf{Dynamic Batching:} Candidate entities are grouped into dynamic batches (default size $k=10$) to fit within the optimal context window of the Textual LLM, balancing token consumption with inference latency.
    \item \textbf{Parallelization:} Semantic checks are independent events. A thread pool architecture is utilized to process $(p, \text{batch}_i)$ tasks in parallel, significantly reducing total wall-clock time.
    \item \textbf{Fine-Grained Fault Tolerance:} The system maintains an atomic progress log, recording results at the batch level. This allows for fine-grained checkpoints, enabling the pipeline to resume graph construction from the exact point of interruption without redundant computation.
\end{itemize}

The final output is a NetworkX\footnote{\url{https://networkx.org/en/}} compatible JSON graph, which serves as the navigational structure for the subsequent Blueprint Generation phase, allowing the system to distinguish between \textit{Core Entities} (Actors) and \textit{Peripheral Entities} (Resources).

\subsection{Service Blueprint Generation (Step 4)}
Leveraging the established \textit{Person-Centric Entity Graph}, the subsequent challenge is to instantiate discrete, functional environments termed \textit{Service Blueprints}, which define the interaction space for an LLM agent. A Blueprint effectively formalizes an MCP Server, encapsulating the state schema, data relationships, and executable tools available to a specific user role.

We propose a multi-stage \textbf{Coarse-to-Fine Generation Framework} to synthesize these blueprints. Formally, a Blueprint $\mathcal{B}$ is defined as a tuple $\langle \mathcal{C}, \mathcal{P}, \mathcal{R}, \mathcal{F} \rangle$, where $\mathcal{C}$ is the Core Entity (Actor), $\mathcal{P}$ is a set of Peripheral Entities (Resources), $\mathcal{R}$ is a set of relational data schemas, and $\mathcal{F}$ is a set of executable functions.

\subsubsection{Stage 1: Topology Sampling (Outline Generation)}
The first stage focuses on structural diversity. For every Person node identified in the Entity Graph, the system generates $k$ distinct \textit{Blueprint Outlines}. An outline represents a potential use-case scenario (e.g., ``CourseEnrollmentManager''). The system samples a subgraph centered on the Core Entity $\mathcal{C}$ and selects $1 \dots N$ Peripheral Entities $\mathcal{P}$ from its immediate neighbors. This step ensures that the resulting service boundaries are logically coherent and grounded in the domain topology.

\subsubsection{Stage 2: Functional Synthesis with Feedback Loop}
This stage expands the high-level outline into a detailed specification. The Textual LLM is tasked with defining:
\begin{itemize}
    \item \textbf{Relational Schemas ($\mathcal{R}$):} Defining the data tables that track state (e.g., \textit{StudentCourseEnrollment}). To ensure database integrity, a strict \textbf{Star-Schema Constraint} is enforced: every relationship must include the Core Entity as a foreign key, and relationships cannot reference other relationships.
    \item \textbf{Tool Definitions ($\mathcal{F}$):} Generating API signatures (e.g., \texttt{enroll\_in\_course}) that agents invoke to manipulate the state.
\end{itemize}

To mitigate the stochastic nature of LLMs, a \textbf{Constraint-Aware Feedback Mechanism} is implemented. The generated blueprint is validated against rigid structural rules (e.g., maximum relationship depth, mandatory authorization functions). If a violation is detected (e.g., a schema missing the Core Entity reference), the system constructs a feedback prompt containing the specific error log and the rejected draft, instructing the Textual LLM to iteratively refine the output. This loop repeats up to $M$ times, significantly increasing the yield of compilable blueprints.

\subsubsection{Stage 3 \& 4: Semantic Refinement and Syntactic Sanitization}
The final phases focus on consistency and execution safety. A \textit{Fixup} pass utilizes the Textual LLM to align terminologies, correcting entity name hallucinations (e.g., aligning ``Advisor'' to the canonical ``AcademicAdvisor''). Finally, a deterministic \textit{Local Validation} pass sanitizes identifiers to ensure they are valid Python tokens and normalizes foreign keys to follow the strictly typed format $\{entity\_name\}\_id$. This ensures that the generated Blueprints are not only semantically valid but also syntactically executable in the downstream environment deployment.

\subsection{Tool Interface Standardization (Step 5)}
To operationalize the semantically generated Blueprints, the system enters the \textit{Tool List Formulation} phase. This module serves as a deterministic interface adapter, translating the abstract functional definitions $\mathcal{F}$ within each Blueprint into the standardized JSON Schema specification required by modern LLMs (specifically the OpenAI Function Call format). This step ensures that the synthesized services are strictly machine-readable and executable by autonomous agents.

\subsubsection{Schema Transformation Logic}
The transformation process $\Phi$ maps a Blueprint function definition $f_{bp}$ to an executable tool definition $T_{schema}$. Formally, for each function $f \in \mathcal{F}$, the transformation is defined as:
$$ T_{schema} = \Phi(f_{name}, f_{desc}, f_{params}) $$
where the internal parameter types (e.g., Python-style \texttt{str}, \texttt{int}) are coerced into JSON Schema types (e.g., \texttt{string}, \texttt{integer}).

To ensure the robustness of tool invocation during agent interaction, a \textbf{Strict Parameter Requirement Policy} is implemented. Unlike standard programming interfaces where arguments may be optional, our formulation automatically populates the \texttt{required} field of the JSON Schema with \textit{all} defined parameters. This constraint minimizes the probability of agents hallucinating default values or omitting critical arguments during the inference process.

\subsubsection{Modular Serialization}
To support scalable distributed deployment, the system serializes the tool definitions for each MCP Server into isolated JSON artifacts (e.g., \texttt{RetailCatalogCheckout.json}). This modular output structure allows downstream agent environments to dynamically load specific toolsets on-demand, maintaining a clean separation of concerns between the service definition layer and the agent execution runtime. The processing is parallelized at the Blueprint level, utilizing a retry-backed execution engine to ensure data integrity during the serialization of large-scale service registries.

\subsection{Constraint-Satisfying Database Synthesis (Step 6)}
\label{sec:single-database}
To ground the generated Blueprints in a realistic operational context, the system must populate the abstract schemas with concrete, semantically consistent instance data. A multi-stage \textit{Database Generation Pipeline} is implemented to synthesize two categories of data: \textit{Entity Instances} (representing the state of individual objects) and \textit{Relationship Instances} (representing the interactions between objects).

The main challenge in synthetic database generation is maintaining relational integrity, ensuring that generated interactions respect the latent logic of the domain (e.g., a math teacher should not be assigned to teach a history course). To address this, we propose a Summary Constraint Generation Architecture.

\subsubsection{Entity Instantiation and Summarization}
In the first phase, the system generates independent entity tables via Coding LLM-synthesized Python scripts, leveraging libraries such as \texttt{Faker}\footnote{\url{https://faker.readthedocs.io/en/master/}} to produce high-diversity attributes (e.g., names, timestamps, UUIDs). Crucially, immediately following generation, an \textit{Automated Summarization Module} analyzes the output distribution. It extracts precise value ranges, categorical distributions, and data types for every attribute, producing a structured meta-profile (e.g., noting that \texttt{Teacher.specialty} contains values $\{\text{Math}, \text{Physics}\}$).

\subsubsection{Constraint-Driven Relationship Construction}
Naive random sampling frequently results in semantic violations in relational data. We introduce a \textit{Constraint Identification} sub-step that utilizes the entity summaries to pre-compute logic rules. The Textual LLM analyzes the attribute profiles of participating entities to detect potential conflicts, such as:
\begin{itemize}
    \item \textbf{Semantic Mismatch:} Attributes that must align across entities (e.g., \texttt{Doctor.department} == \texttt{Appointment.department}).
    \item \textbf{Temporal Consistency:} Logic governing time-series data (e.g., \texttt{start\_time} $<$ \texttt{end\_time}).
\end{itemize}

Based on these identified constraints, the system generates \textit{Active Construction Scripts}. Instead of inefficient ``sample-and-filter'' rejection sampling, which has $O(N \times M)$ worst-case complexity, these scripts employ an \textbf{Index-Based Sampling Strategy}. The generator pre-indexes candidate entities by their constraint keys (e.g., grouping courses by subject), allowing for $O(1)$ retrieval of valid pairs. This ensures that the resulting relational database is not only syntactically correct but also semantically plausible, mimicking real-world operational distributions.

\subsubsection{Self-Healing Code Execution}
Given the stochastic nature of code generation, syntax errors are inevitable. We integrate a \textbf{Self-Healing Execution Sandbox}. If a data generation script fails during execution, the error trace is captured and fed back into a \textit{Block Editor} (Coding LLM), which iteratively patches the code. This mechanism, combined with fine-grained checkpointing, ensures pipeline robustness even when generating complex, large-scale datasets.

\subsection{Domain Policy Generation (Step 7)}
While the Blueprint defines the \textit{structure} of a service (i.e., existing tools), it lacks the \textit{logic} of operation (i.e., success conditions). To close this gap, the system advances to the \textit{Policy Generation} phase. This module synthesizes a comprehensive \textit{Domain Policy Document} for each MCP Server, establishing the deterministic ground truth that governs agent-system interactions.

\subsubsection{Policy as a Deterministic Contract}
In this framework, a Policy is not merely a user manual but a computational contract. It rigorously defines the execution logic for every tool $t \in \mathcal{F}$, covering four dimensions:
\begin{enumerate}
    \item \textbf{Preconditions:} System states that must hold true before invocation (e.g., ``User session must be authenticated'').
    \item \textbf{Input Validation:} Constraints on parameter formats (e.g., ``\texttt{user\_id} must be a valid UUID v4'').
    \item \textbf{Permission Outcomes:} A decision tree mapping state variables to access decisions (e.g., ``Permitted if \texttt{is\_active} is true; otherwise Rejected'').
    \item \textbf{Side Effects:} The exact state mutations triggered by successful execution.
\end{enumerate}

This structured approach ensures that the environment responds deterministically to agent actions, eliminating the ambiguity often found in purely natural-language-based simulators.

\subsubsection{Closed-World Schema Adherence}
A critical challenge in generating synthetic policies is hallucination—\textit{Textual LLMs} often invent business rules referencing non-existent data attributes (e.g., checking a \texttt{premium\_status} flag that does not exist in the database). To mitigate this, a \textbf{Closed-World Constraint} is enforced. The generating Textual LLM is fed the \textit{Database Summary} from Section~\ref{sec:single-database} as a strict boundary condition. It is explicitly forbidden from referencing any attribute or relationship not present in the generated schema.

To enforce this constraint computationally, a \textbf{Policy Auditor} module is introduced. This secondary Textual LLM parses the generated policy against the database schema, detecting hallucinations such as fabricated attributes or timeline violations (e.g., using real-world time instead of simulation time). If violations are found, the Auditor generates \textit{Search-and-Replace} patches to automatically align the policy with the ground-truth schema.

\subsubsection{Structured Artifacts for Downstream Consumption}
To facilitate automated parsing in later stages (such as \textit{Trajectory Generation} and \textit{Instance Selection}), the policy documents are encapsulated with machine-readable XML-style delimiters. This hybrid format combines the semantic richness of natural language (for agent comprehension) with the structural rigidity of code (for system execution), serving as the central reference point for the entire simulation lifecycle.

\subsection{Policy-Compliant Executable Synthesis (Step 8)}
Upon defining the behavioral rules (Policies) and static tool definitions, the system proceeds to instantiate the runtime environment. This phase, \textit{MCP Server Implementation}, synthesizes fully functional Python code for each MCP server. Unlike standard code generation tasks that rely on open-ended prompts, this approach is strictly grounded in the formal specifications derived from previous steps, ensuring that the generated executable is a faithful implementation of the Domain Policy.

\subsubsection{Template-Guided Code Generation}
To ensure architectural consistency across hundreds of generated services, a \textbf{Template-Filling Strategy} is employed. The Coding LLM is provided with a skeletal Python framework that defines the mandatory class structure, session management logic, and a unified invocation router. The model's task is to populate this skeleton with domain-specific logic, translating the natural language constraints from the Policy Document into executable Python assertions.

A critical feature of the implementation is the \textbf{Confirmation Pattern} for state-mutating operations. For any tool $f$ that modifies the database (e.g., \texttt{create\_order}), the generated code automatically injects a \texttt{confirm} parameter. This enforces a ``Preview-then-Commit'' workflow, allowing the agent to validate the effects of an action before final execution, thereby increasing safety in autonomous operations.

\subsubsection{Feedback-Based Self-Repair}
Code generation is inherently prone to subtle logical errors. To address this, the \textbf{Test-Driven Self-Correction} mechanism is implemented. Immediately after generating the server code, the system synthesizes a corresponding test suite using \texttt{pytest}, covering unit tests, session isolation, and policy violation scenarios.

The Test-Driven Self-Correction loop operates as follows:
\begin{enumerate}
    \item \textbf{Execution:} The system runs the generated test suite against the server implementation.
    \item \textbf{Diagnosis:} If a test fails, the error trace and the relevant source code are fed into a \textit{Diagnostic Coding LLM}. This model classifies the failure into one of three categories: \textit{Implementation Bug} (logic error in server), \textit{Test Defect} (error in test case), or \textit{Data Issue} (invalid test data).
    \item \textbf{Patching:} Based on the diagnosis, a specialized \textit{Block Editor} (Coding LLM) generates a precise ``Search-and-Replace'' patch to fix the identified file.
    \item \textbf{Verification:} The tests are re-run. This cycle repeats up to $K$ times (default $K=5$) until all tests pass or the budget is exhausted.
\end{enumerate}

This closed-loop system ensures that the final deployed servers are not only syntactically correct but also functionally robust and compliant with the logical constraints of the domain.

\subsection{Semantic Tool Dependency Modeling \& Task Generation (Steps 9-10)}
Although the previous step establishes the executable implementation of tools, it lacks the temporal and logical context required for coherent agent execution. A flat list of executable functions (e.g., \texttt{login}, \texttt{buy\_item}, \texttt{logout}) does not inform an agent about the necessary order of operations. To resolve this, the \textit{Tool Graph Generation} phase structures the functional space into a DAG $G_{tool} = (\mathcal{V}_t, \mathcal{E}_t)$, where nodes represent executable tools and edges represent execution dependencies, serving as the foundation for downstream Task Template generation.

\subsubsection{Dependency Taxonomy}
Three distinct categories of edges in $\mathcal{E}_t$ are defined, prioritized by their constraint rigidity:
\begin{enumerate}
    \item \textbf{State Dependencies (Hard Constraint):} Operations that require a specific system state prerequisite. The most prominent example is the \textit{Authentication Gate}, where sensitive operations depend on an active session state established by an \texttt{authorize} call.
    \item \textbf{Information Flow (Data Constraint):} Dependencies where the output artifact of tool $A$ (e.g., a \texttt{customer\_id} returned by search) serves as a mandatory input parameter for tool $B$ (e.g., \texttt{get\_profile}).
    \item \textbf{Storyline Flow (Soft Constraint):} Probabilistic transitions based on typical user behavior patterns (e.g., \texttt{view\_item} $\rightarrow$ \texttt{add\_to\_cart}), used to guide the graph connectivity when no hard constraints exist.
\end{enumerate}

\subsubsection{Topological Constraints for Valid Trajectories}
To ensure that any traversal of $G_{tool}$ yields a valid execution trace, strict topological properties are enforced during generation:
\begin{itemize}
    \item \textbf{Single Source of Truth:} The graph must possess exactly one node with an in-degree of zero ($d_{in}(v)=0$), typically the \texttt{authorize} function. This guarantees that all generated trajectories inevitably begin with a valid authentication handshake.
    \item \textbf{Full Reachability:} The graph must be strongly connected in the forward direction from the source node. Formally, for every node $v \in \mathcal{V}_t$, there must exist a path $p: v_{source} \leadsto v$. This ensures that no tool is functionally unreachable in the simulation.
    \item \textbf{Acyclicity:} The graph must be a DAG to prevent infinite loops in automated planning generation.
\end{itemize}

\subsubsection{Generation and Validation}
The construction of $G_{tool}$ is performed by the Textual LLM reasoning over the Domain Policy and the implemented tool interfaces. The model explicitly identifies the dependency type for each edge. Following generation, a deterministic graph validation module audits the topology. If a graph violates any topological constraint (e.g., presence of a cycle or a disconnected component), the system triggers a regeneration cycle with feedback, ensuring that the downstream task generator receives only logically sound workflow templates.

\subsection{Semantic-Driven Cross-Domain Trajectory Fusion (Step 11)}
\label{sec:trajectory_fusion}
While single domains offer atomic capabilities, complex real-world tasks often span multiple service boundaries (e.g., booking a medical appointment and immediately filing an insurance claim). To simulate such high-fidelity scenarios, the \textit{Trajectory Fusion} module is introduced.

\begin{figure*}[!t]
\centering
    \includegraphics[width=0.8\linewidth]{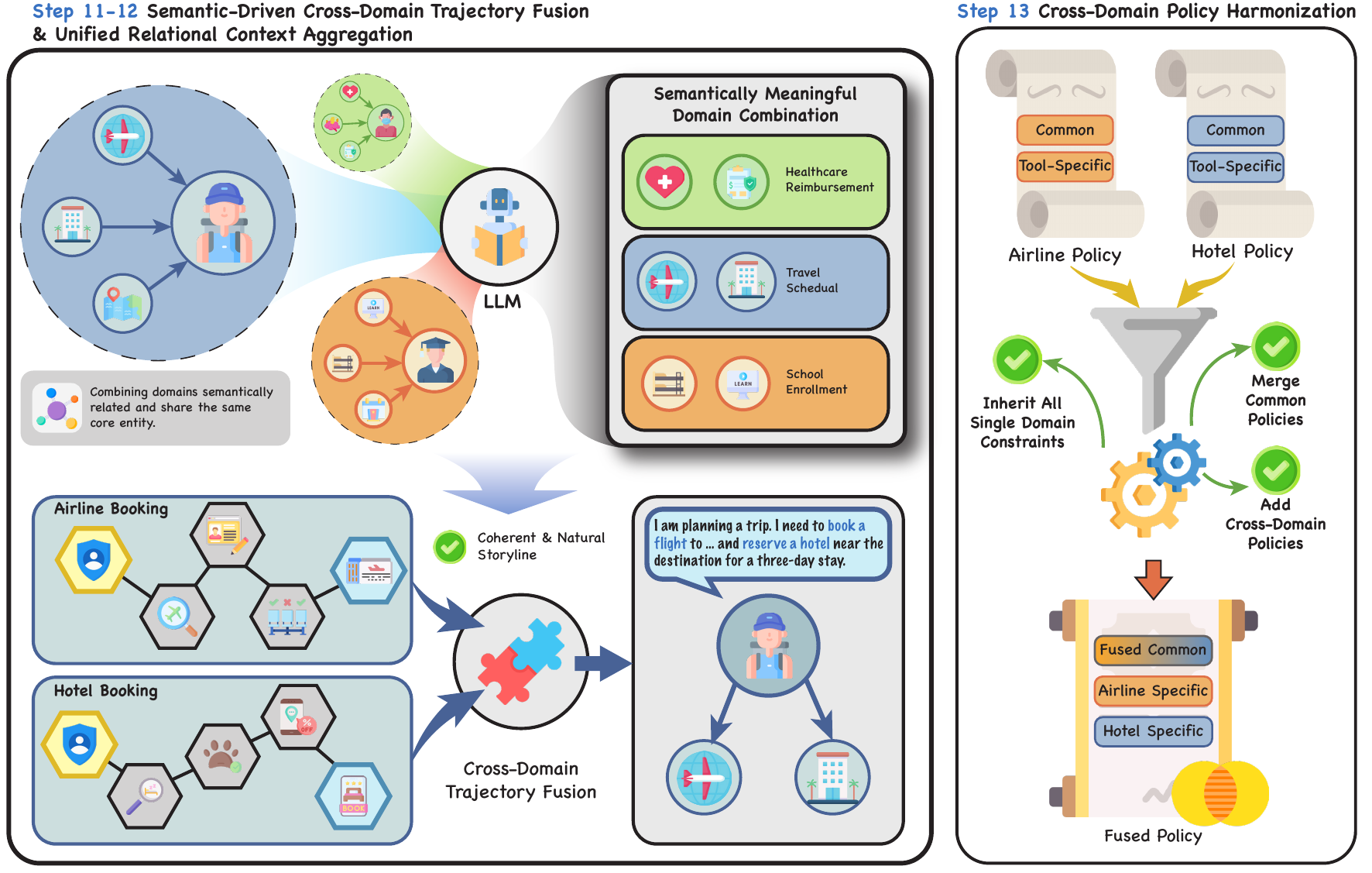} % 替换为你的文件名
    \caption{\textbf{The Semantic Cross-Domain Fusion Mechanism.} Building upon Figure~\ref{fig:single_domain}, this phase (Steps 11--13) synthesizes composite scenarios. The left panel illustrates \textit{Trajectory Fusion} via shared entities, while the right panel depicts \textit{Policy Harmonization} to merge conflicting rules into a unified governance policy.}
    \label{fig:cross_domain}
\end{figure*}

As detailed in Figure~\ref{fig:cross_domain}, this phase represents the \textbf{Cross-Domain Composition Layer}, which builds directly upon the atomic, validated environments established in Figure~\ref{fig:single_domain}. The mechanism operates on two levels:
\begin{itemize}
    \item \textbf{Trajectory Interlocking (Left Panel):} The system semantically links distinct workflows (e.g., Airline and Hotel) via shared core entities, synthesizing coherent storylines that require multi-hop reasoning.
    \item \textbf{Policy Harmonization (Right Panel):} To ensure governance consistency, conflicting rules from different domains are merged through a synthesis funnel (Step 13), producing a unified policy that governs the composite interaction.
\end{itemize}
This hierarchical approach allows AgentSkiller to scale complexity without compromising the execution stability guaranteed by the single-domain pipeline.

\subsubsection{Feasibility-Aware Fusion Logic}
Let $\mathcal{G}_A$ and $\mathcal{G}_B$ be the tool execution graphs for two distinct domains $D_A$ and $D_B$, identified as a viable pair in the previous \textit{Domain Combination} step based on their semantic relationship. A cross-domain trajectory is defined as a tuple $\tau_{cross} = \langle \tau_A, \tau_B, M \rangle$, where $\tau_A$ and $\tau_B$ are tool sequences from respective domains, and $M$ is a natural language \textit{Motivation Descriptor}.

A critical innovation in the framework is \textbf{Single Domain Feasibility Filtering}. Cross-domain trajectories are essentially composed of single-domain segments. If a segment (e.g., $\tau_A$) has already proven infeasible during the single-domain validation phase (due to data sparsity or policy conflicts), any cross-domain trajectory containing it is predetermined to fail. Therefore, immediately after the Textual LLM fuses potential trajectories, the system cross-references them against the \textit{Validated Task Registry} produced by the Task Filtering module (see Section~\ref{sec:task_filtering}) from the single-domain pipeline. Trajectories containing unvalidated segments are pruned \textit{ex ante}. This optimization significantly reduces computational waste by preventing the execution of doomed trajectories.

\subsubsection{Incremental Parallel Synthesis}
Given the combinatorial explosion of potential trajectory pairs, efficiency is paramount. We implement a \textit{Domain Combination Level Parallelism} strategy. Each domain pair is processed independently with automated retries for fault tolerance. Crucially, the system employs an \textit{Incremental State Tracker} based on file-system artifacts. This allows the pipeline to support ``resume-from-break-point'' functionality, skipping already fused combinations and focusing solely on pending pairs, thereby ensuring scalability for large-scale benchmarks.

\subsection{Unified Relational Context Aggregation (Steps 12)}
Subsequent to synthesizing cross-domain workflows, the system must establish a unified data environment to support the execution of these composite tasks. Since single-domain databases were generated in Section~\ref{sec:single-database}, simply referencing them individually is insufficient for modeling complex interactions where entities (e.g., a \textit{User}) act as shared state carriers across boundaries. The \textit{Database Fusion} module is implemented to construct a cohesive relational context for each domain pair.

\subsubsection{Entity Unification Strategy}
Let $\mathcal{C}_{cross}$ be a cross-domain combination involving server set $S = \{s_1, s_2, \dots, s_k\}$. The fusion process aggregates the complete entity set $\mathcal{E}_{unified} = \bigcup_{s \in S} \mathcal{E}_s$, where $\mathcal{E}_s$ includes both the core actor and all peripheral resources defined in the blueprint of server $s$. This aggregation is purely deterministic, performed via a high-throughput file-system replication mechanism that copies the relevant JSON table artifacts into a dedicated fusion namespace (e.g., \texttt{Hospital\_Insurance/}). This approach ensures that the subsequent task instantiation phase has immediate access to the full transitive closure of entities required for any multi-hop trajectory.

\subsubsection{Namespace-Isolated Relationship Mapping}
While entities are unified, relationships must maintain their domain specificity to avoid schema collisions (e.g., both domains might define a generic \texttt{UserLog} table). To resolve this, a \textbf{Namespace Isolation Policy} is enforced during aggregation. Every relationship table $R$ originating from server $s$ is renamed to $s\_R$ within the fused context. This simple yet effective transformation preserves the provenance of relational data, allowing the system to verify constraints (e.g., ensuring a \textit{Patient} exists in both the \textit{Appointment} and \textit{Insurance} registries) without ambiguity.

\subsection{Cross-Domain Policy Harmonization (Step 13)}
Similar to database schema unification, the behavioral rules governing tool usage must be harmonized. Simply concatenating the policy documents of two domains (e.g., $P_A$ and $P_B$) is insufficient, as it fails to address inter-domain dependencies or conflicts (e.g., a ``strict privacy'' rule in $D_A$ conflicting with a ``data sharing'' requirement in $D_B$). To resolve this, the \textit{Policy Merge} phase synthesizes a unified governance document $P_{unified}$.

\subsubsection{Conflict Resolution and Rule Synthesis}
The system employs a Textual LLM-based mediator to ingest the discrete policy sets. Its primary objective is not merely aggregation but \textit{harmonization}. The model is tasked with:
\begin{enumerate}
    \item \textbf{Conflict Detection:} Identifying and reconciling contradictory constraints between domains.
    \item \textbf{Interaction Rule Synthesis:} Generating new ``bridge rules'' that govern the interface between domains. For example, if Domain A is a booking system and Domain B is a payment gateway, the synthesizer might add a rule: \textit{``Payment (Domain B) must only be initiated after a Booking (Domain A) reaches 'confirmed' status.''}
\end{enumerate}

\subsubsection{Structured Policy Preservation}
To ensure that the downstream planner can accurately parse tool-specific constraints, the merger strictly preserves the structural integrity of the input policies. Using a marker-based strategy, the system ensures that while global rules are merged, the precise preconditions and post-conditions for individual tools remain isolated and identifiable, preventing ``hallucinated'' constraints from leaking between unrelated functions.

\subsection{Task Template Synthesis via Trajectory Enumeration (Step 14)}
While the Tool Graph provides the theoretical execution space for an agent, it does not directly yield user instructions. To train or evaluate agents, these graph topologies must be converted into actionable natural language prompts. This phase, \textit{Task Template Generation}, automates the creation of high-quality task descriptions by mapping valid execution paths to semantic user intents.

\subsubsection{Trajectory Enumeration}
We first identify valid execution sequences, termed \textit{trajectories}. For single-domain tasks, the system systematically explores the tool graph $G_{tool}$ via a path-finding algorithm. For cross-domain scenarios, the composite trajectories synthesized in Section~\ref{sec:trajectory_fusion} are utilized. A trajectory $\tau = [t_1, t_2, \dots, t_k]$ is a linear path of tool nodes such that:
\begin{enumerate}
    \item $t_1$ is a valid entry point (typically an authorization gate, where $d_{in}(t_1)=0$).
    \item Each transition $(t_i, t_{i+1})$ corresponds to a valid edge in $\mathcal{E}_t$.
    \item The sequence terminates at a logical endpoint ($d_{out}(t_k)=0$) or reaches a predefined maximum length $L_{max}$.
\end{enumerate}
A path-finding algorithm (based on NetworkX) is employed to enumerate all simple paths satisfying $L_{min} \le |\tau| \le L_{max}$, ensuring a diverse dataset of task complexities ranging from simple lookups to multi-step workflows.

For single-domain tasks, the generated instance combinations are forwarded to the subsequent \textit{Task Filtering} step for execution validation. In contrast, for cross-domain scenarios, validation is integrated directly into the creation phase (Mode 2) to handle the sparsity of valid combinations efficiently.

\subsubsection{Semantic Template Generation}
For each trajectory, the Textual LLM is tasked with synthesizing a corresponding \textit{Task Template}. The model receives the sequence of tools, their associated policy constraints, and the database schema. Its objective is to generate a natural language instruction that:
\begin{itemize}
    \item \textbf{Encodes Motivation:} Explains \textit{why} the user wants to perform this sequence (e.g., ``I need to register for a class'').
    \item \textbf{Preserves Variables:} Uses symbolic placeholders (e.g., \texttt{<Student.id>}) instead of hardcoded values, allowing for dynamic instantiation in later stages.
    \item \textbf{Avoids Hallucination:} Strictly adheres to the functionality provided by the tools in $\tau$.
\end{itemize}

\subsubsection{Automated Quality Assurance}
To filter out low-quality generations (e.g., instructions that are logically incoherent or fail to trigger all tools in the trajectory), a distinct \textit{LLM-as-a-Judge} module is integrated. This evaluator scores each generated template on a 5-point scale across five dimensions: Semantic Clarity, Logical Coherence, Completeness, Naturalness, and Specificity. Only templates exceeding an aggregate threshold $\theta$ are retained in the final dataset, ensuring a high-fidelity training corpus.

\subsection{Constraint-Aware Instance Grounding (Step 15)}
\label{sec:constraint-grounding}
A primary limitation in current tool-use benchmarks is the decoupling of instructions from the underlying database state. An instruction like ``Refund the order'' is only executable if there exists an order in the database with \texttt{status=`refundable'}. To bridge this gap, the \textit{Instance Combos Selection} phase is introduced. This phase grounds abstract Task Templates into concrete database instances using a \textbf{Dual-Mode Strategy} tailored to domain complexity: \textit{Sampling} for single domains and \textit{Creation} for cross-domain tasks.

\subsubsection{Mode 1: Single Domain Sampling}
For single-domain tasks, where data density is high, a \textbf{Plan-Execution Architecture} is employed:
\begin{enumerate}
    \item \textbf{Phase 1: Constraint Analysis (Plan):} The Textual LLM analyzes the trajectory and Domain Policy to identify \textit{Trajectory Blocking Values}—terminal states that would render subsequent tools inoperable (e.g., selecting a 'shipped' order for a 'cancel' tool). It outputs a structured analysis comprising relationship operation types (CREATE/READ/UPDATE) and sequential constraints.
    \item \textbf{Phase 2: Code Synthesis (Execute):} Based on the plan, the Coding LLM synthesizes a Python script to sample instances. It utilizes \textit{Relationship-First Sampling}, prioritizing sampling from relationship tables (e.g., \texttt{Enrollments}) over independent entity tables to guarantee referential integrity.
\end{enumerate}

\subsubsection{Mode 2: Cross-Domain Creation and Validation}
In cross-domain scenarios, finding pre-existing instances that satisfy the intersection of multiple domain policies is statistically sparse. Therefore, a \textbf{Creation-Validation Mode} is adopted:
\begin{enumerate}
    \item \textbf{Dependency Analysis \& Creation:} The system parses explicit placeholders and infers implicit dependencies, allowing the Coding LLM to synthesize a \textit{Creation Code}. This script generates new, policy-compliant entity and relationship instances specifically tailored to the trajectory's constraints.
    \item \textbf{Parameter Completeness Verification:} Before execution, the system verifies that the generated `value\_domain\_samples` cover every parameter of every tool in the trajectory. This step also enforces a consistent temporal anchor (e.g., fixing ``now'' to a static timestamp) to prevent temporal drift during execution.
    \item \textbf{Transactional Validation:} To ensure isolation, the system performs a database backup before writing the new instances. A \textit{Trajectory Executor} then runs the task against the augmented database. If successful, the data is committed and the trajectory is serialized into the \textit{Validated Task Registry}; if failed, the database is rolled back, and a \textit{Block Editor (Coding LLM)} iteratively fixes the creation code based on error logs.
\end{enumerate}

This dual approach ensures that AgentSkiller efficiently handles both high-density single-domain tasks and complex, sparse cross-domain interactions without compromising database integrity.

\subsection{Execution-Based Task Filtering (Step 16)}
\label{sec:task_filtering}

Targeting primarily the single-domain trajectories generated via the Sampling Mode, this step functions as a dynamic verification layer. Since the instance combinations generated in the previous step are theoretically derived from static constraints, they may still fail due to implicit logic gaps or runtime exceptions. To ensure the reliability of the benchmark, the \textit{Task Filtering} module acts as a dynamic verification layer. This step explicitly executes the generated trajectories on the MCP servers using the synthesized instance data, filtering out any combinations that fail to reach the intended terminal state. (Note: Cross-domain trajectories bypass this step as they undergo intrinsic validation during their creation in Section~\ref{sec:constraint-grounding}.)

\paragraph{Dual-Path Execution Mechanism.}
Validating thousands of trajectories via LLM-driven execution is computationally expensive. To address this, a \textbf{Dual-Path Execution Mechanism} is implemented to optimize resource utilization:
\begin{itemize}
    \item \textbf{Slow Path (Generative Execution):} For a novel trajectory $\tau$ without existing cached code, the Coding LLM synthesizes a dedicated Python execution script using the \texttt{TrajectoryExecutor}. This script dynamically handles authentication, tool invocation sequences, and return value assertions.
    \item \textbf{Fast Path (Cached Execution):} Once a script is successfully verified for a trajectory template, it is hashed and cached. Subsequent validations for the same trajectory structure (but different data instances) bypass the LLM generation phase and directly execute the cached code.
\end{itemize}
This mechanism reduces token consumption by orders of magnitude while enabling massive parallel processing.

\paragraph{Root Cause Analysis and Filtering.}
To ensure the robustness of the final dataset, the system employs a trajectory-wise parallel processing strategy. When an execution fails, the system does not simply discard the data but performs an automated \textit{Error Taxonomy Analysis}. Failures are classified into \textit{Policy Violations}, \textit{Invalid Entity Data}, or \textit{Server Implementation Bugs}. Only instance combinations that pass execution without error are serialized into the \textit{Validated Task Registry}, ensuring that the downstream evaluation dataset is 100\% executable and bug-free.

\subsection{Task Instantiation with Scenario Grounding (Step 17)}
The final stage of the generation pipeline transforms the execution-validated tuples (derived from the \textit{Validated Task Registry}), comprising the tool trajectory $\tau$ and the validated database instance $C_i$ into a naturalistic interaction interface suitable for agent evaluation. This phase, \textit{Task Instantiation}, bridges the gap between the structured ``System View'' (database records, function calls) and the unstructured ``User View'' (intent, ambiguity, partial information).

\subsubsection{Persona-Based Query Synthesis}
Naive template filling (e.g., simply replacing placeholders in ``I want to delete order \texttt{<id>}'') often results in robotic, overly explicit prompts that fail to test the conversational capabilities of an agent. To address this, a \textbf{Persona-Based User Simulator} is employed. The Textual LLM is prompted to adopt the persona of a user facing a specific problem grounded in the instance data $C_i$.

The simulator is tasked with generating three distinct artifacts:
\begin{enumerate}
    \item \textbf{The Grounded Instruction:} A fully instantiated version of the task template where variables are replaced with concrete values (e.g., ``Grant John Smith access...''), serving as the gold-standard reference.
    \item \textbf{The Motivation Context:} A plausible backstory explaining \textit{why} the user is performing this action (e.g., ``I have a client meeting in an hour''), enriching the semantic context.
    \item \textbf{The Startup Query:} The initial natural language utterance sent to the agent.
\end{enumerate}

\subsubsection{Information Asymmetry and Ambiguity}
A key contribution of the query generation strategy is the enforcement of the \textbf{Information Hiding Principle}. Real-world users rarely provide all necessary parameters in the first turn. To simulate this, the Startup Query is deliberately constrained to focus on a ``Single Immediate Pain Point'' (e.g., ``I'm having trouble with my building access'') rather than a complete specification. This forces the evaluating agent to demonstrate \textit{active information gathering} skills—asking clarifying questions to retrieve the necessary entities (e.g., ``Which door?'' or ``Who is the visitor?'')—rather than merely acting as a slot-filling engine.

\subsubsection{Standardized Evaluation Artifacts}
The output of this phase is serialized into the JSON Lines (JSONL) format, creating a standardized benchmark dataset. Each record encapsulates the user's initial prompt, a hidden ``User System Prompt'' containing the private knowledge required to answer the agent's questions, and the ground-truth trajectory hash. This structure enables automated evaluation frameworks to not only check the final execution success but also assess the efficiency of the multi-turn dialogue.

\subsection{Multi-Turn Dialogue Rollout}
\label{sec:rollout}

To conclude the data synthesis, the Rollout step is executed. This phase transitions from static task definitions to dynamic interaction logs by orchestrating a closed-loop dialogue between a target \textit{Agent} and a \textit{User Simulator} within the instantiated environment.

Initialized with the \textit{Startup Query} and the hidden \textit{User System Prompt} generated in Step 17, the User Simulator initiates the conversation to achieve its specific persona-based goal. The system acts as an orchestrator, forwarding the natural language input from the user to the Agent and executing the Agent's tool calls against the mock MCP servers. This interactive loop continues until the Simulator determines that the latent goal has been satisfied or a termination condition is met. The result is a set of high-fidelity, multi-turn trajectories that capture the complete interplay between agent reasoning, tool execution, and environment feedback.

\subsection{Multi-Dimensional Evaluation}
\label{sec:evaluation}

To rigorously assess the data quality of the dataset generated via AgentSkiller, a comprehensive evaluation system is introduced, specifically designed for long-horizon, multi-turn tool-calling tasks. Unlike traditional benchmarks that rely solely on text matching, a \textit{Composite Evaluator} architecture is implemented to assess both objective correctness (action execution and state transitions) and subjective interaction quality.

\subsubsection{Scoring Philosophy: The ``All-or-Nothing'' Metric}
This framework is underpinned by a strict ``all-or-nothing'' scoring paradigm. The success of a generalist agent is defined not by partial completion, but by its ability to execute a trajectory flawlessly across all dimensions. The final evaluation score $S_{final}$ is calculated as the product of individual evaluator scores:

\begin{equation}
 S_{final} = \prod_{i \in \mathcal{E}} s_i, \quad \text{where } s_i \in \{0.0, 1.0\}
\end{equation}

Here, $\mathcal{E}$ represents the set of enabled evaluators (Action, Environment, Subjective). Consequently, a failure in any single dimension—whether it be a missed tool call, a database consistency error, or a hallucination—results in a global failure for the task.

\subsubsection{Trajectory Pre-processing: Intelligent Pruning}
To prevent false negatives arising from efficient but non-standard agent behaviors, a \textit{Trajectory Pruner} is introduced. This module preprocesses the ground-truth ``golden'' trajectories to identify and remove redundant operations before evaluation. The pruning logic adheres to two strict invariants:
\begin{enumerate}
   \item \textbf{State Preservation:} State-changing operations (e.g., \texttt{create}, \texttt{update}) are never pruned to ensure system integrity.
   \item \textbf{Redundancy Elimination:} Read-only operations are pruned if and only if the information they retrieve is strictly a subset of information provided by preceding calls.
\end{enumerate}
For instance, if a \texttt{list\_accounts} call returns balance details for all users, a subsequent \texttt{check\_balance} call for a specific user is deemed redundant and removed from the evaluation requirements.

\subsubsection{Objective Evaluation Components}
The objective evaluation ensures that the agent strictly adheres to the ground-truth logic and maintains system integrity.

\noindent
\textbf{Action Verification.}
This component verifies the sequence of tool invocations by comparing the agent's execution rollout against the pruned golden trajectory. It employs a flexible matching logic that permits:
\begin{itemize}
   \item \textbf{Order Independence:} Allowing valid permutations of independent tool calls.
   \item \textbf{Non-Destructive Additions:} Tolerating extra tool calls (e.g., additional searches) provided they do not negatively impact the state.
   \item \textbf{Fuzzy Parameter Matching:} Enforcing strict adherence for core parameters while allowing tolerance for floating-point values ($\epsilon=1e-4$) and case-insensitivity for string literals. Non-deterministic fields (e.g., free-text filters) are automatically ignored.
\end{itemize}

\noindent
\textbf{Diff-Based State Consistency (EnvironmentEvaluator).}
To verify referential integrity without being brittle to stochastic identifier generation (e.g., UUIDs), a \textbf{Diff-Based Comparison} strategy is implemented. Rather than comparing absolute final states, the state delta induced by the agent ($\Delta_{agent}$) is compared with that of the golden trajectory ($\Delta_{gold}$) relative to the initial state. The success condition is defined as:
\begin{equation}
\Delta_{gold} \subseteq \Delta_{agent}
\end{equation}
This condition ensures that the agent performs all mandatory state mutations (inclusion validity) while permitting additional, valid auxiliary operations. Records are compared by content rather than identity, ensuring robustness against ID variations.

\subsubsection{Subjective Quality and Safety (LLM-as-a-Judge)}
Beyond functional correctness, an LLM-as-a-Judge mechanism is employed to assess the nuance of agent interactions, utilizing a detailed taxonomy of error categories:

\begin{itemize}
   \item \textbf{Hallucination Detection (H1--H5):} We rigorously flag groundedness errors, including status misrepresentation (H1), policy hallucination (H2), and unverified assumption of user intent (H5).
   \item \textbf{Communication Efficiency (R1--R2):} The system penalizes redundant behaviors, such as re-requesting provided slots (R1) or querying information already exposed by previous tool outputs (R2).
   \item \textbf{Safety and Policy Compliance (S1--S2):} We enforce strict adherence to safety protocols, specifically flagging agents that omit mandatory confirmation steps before write operations (S1) or violate domain-specific operational policies (S2).
\end{itemize}

\section{Data Statistics and Analysis}
\label{sec:data_statistics}

To validate the scalability and rigor of our {AgentSkiller} framework, a large-scale corpus of multi-turn interaction trajectories was synthesized. The final dataset consists of approximately 11,522 samples processed through the generation pool. This section analyzes the dataset across four critical dimensions: domain distribution, topological connectivity, semantic complexity, and fine-grained error taxonomy.

\subsection{Domain Distribution and Topology}
Leveraging the \textit{Domain Expansion} (Step 1) and \textit{Service Blueprint Generation} (Step 4) modules, the dataset achieves extensive coverage across high-stakes industrial sectors. 

\begin{figure*}[!t]
    \centering
    % Subfigure 1: Pie Chart
    \begin{subfigure}[b]{0.28\textwidth} % 稍微调宽一点以适应图片
        \centering
        \includegraphics[width=\textwidth]{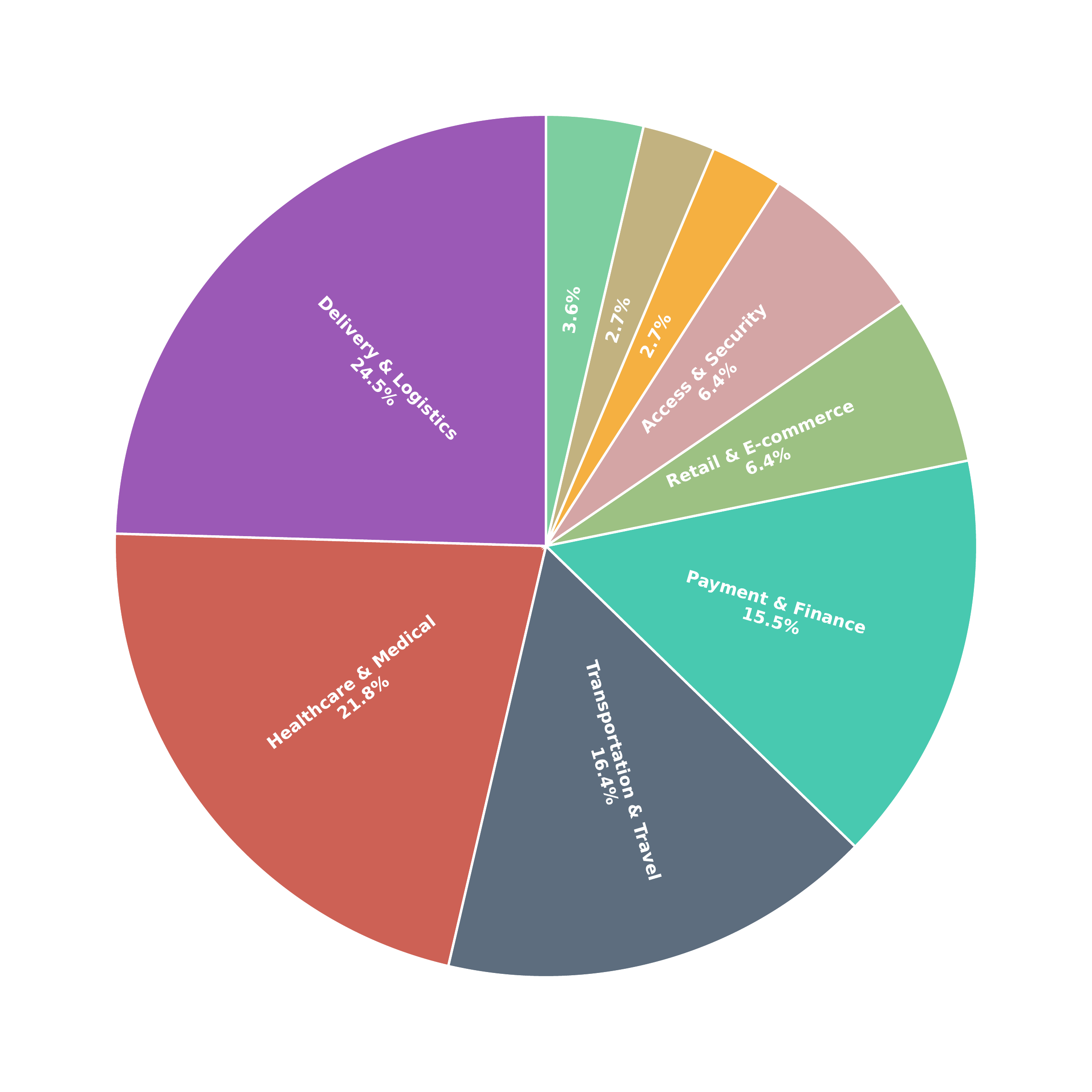} 
        \captionsetup{justification=centering, singlelinecheck=false}
        \caption{Total Data Themes}
        \label{fig:domain_dist}
    \end{subfigure}
    \hfill
    % Subfigure 2: Judge 2 Topology
    \begin{subfigure}[b]{0.70\textwidth}
        \centering
        \includegraphics[width=\textwidth]{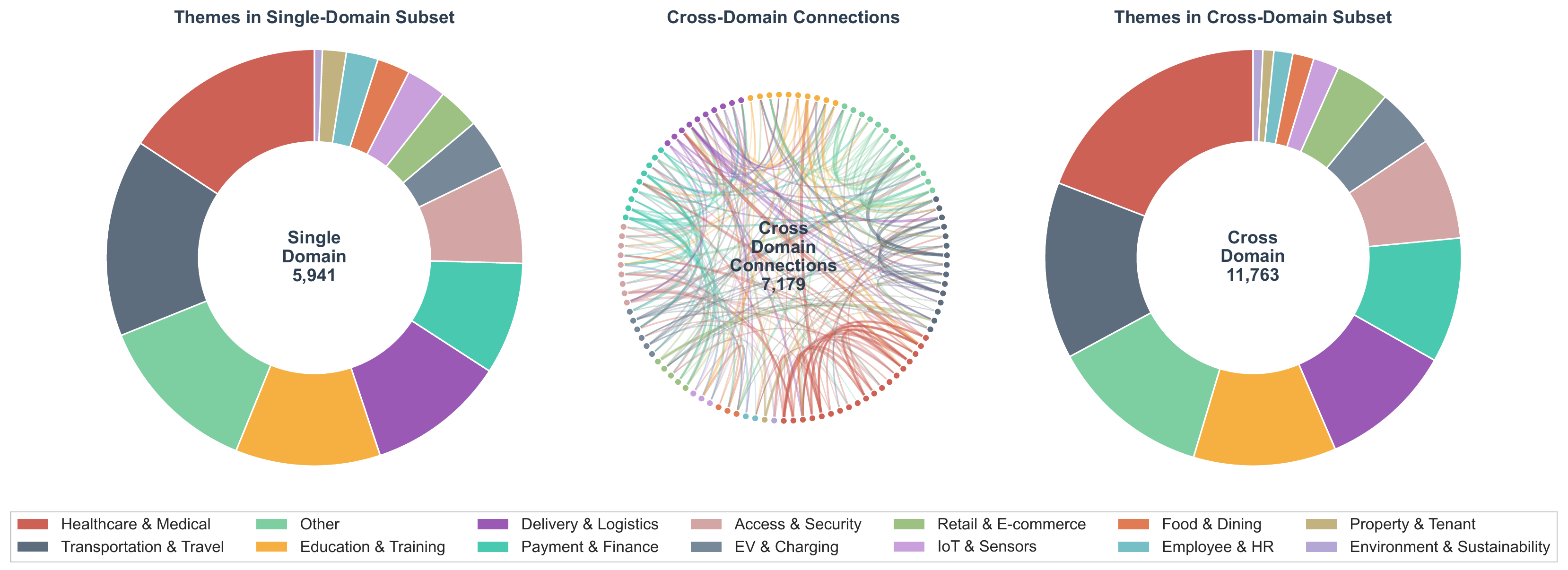} 
        \captionsetup{justification=centering, singlelinecheck=false}
        \caption{Data Themes: Single-Domain vs. Cross-Domain Connectivity}
        \label{fig:domain_topology}
    \end{subfigure}
    \caption{\textbf{Data Theme Analysis.} (a) The dataset prioritizes complex operational topics like Logistics and Healthcare. (b) The framework generates a dense web of cross-domain connections (center), resulting in a Cross-Domain subset (right) that is significantly larger than the Single-Domain subset (left), extending the interaction horizon beyond isolated tasks.}
    \label{fig:domain_analysis}
\end{figure*}

\paragraph{Sector Breakdown.}
As detailed in Figure~\ref{fig:domain_analysis}(a), the synthesized tasks are distributed among complex service domains, with \textbf{Delivery \& Logistics (24.5\%)}, \textbf{Healthcare \& Medical (21.8\%)}, and \textbf{Payment \& Finance (15.5\%)} constituting the majority. These domains were specifically targeted for their rigorous logic constraints and high demand for stateful operations, distinguishing this dataset from general-purpose open-domain benchmarks.

\paragraph{Cross-Domain Connectivity.}
Beyond isolated sectors, Figure~\ref{fig:domain_analysis}(b) visualizes the dense topological connectivity achieved via the \textit{Trajectory Fusion} mechanism (Step 11). The chord diagram demonstrates that the framework successfully bridges distinct service boundaries (e.g., linking \textit{Healthcare} with \textit{Insurance}). Consequently, the volume of tasks in the \textbf{Cross-Domain Subset ($N=11,763$)} significantly surpasses that of the \textbf{Single-Domain Subset ($N=5,941$)}, confirming that AgentSkiller effectively mitigates the scarcity of long-horizon, composite training data.

% --- Figure Group 1: Domain Analysis ---

\subsection{Interaction Complexity and Semantics}
\begin{figure*}[!t]
    \centering
    \begin{subfigure}[b]{0.35\textwidth}
        \centering
        \includegraphics[width=1\textwidth]{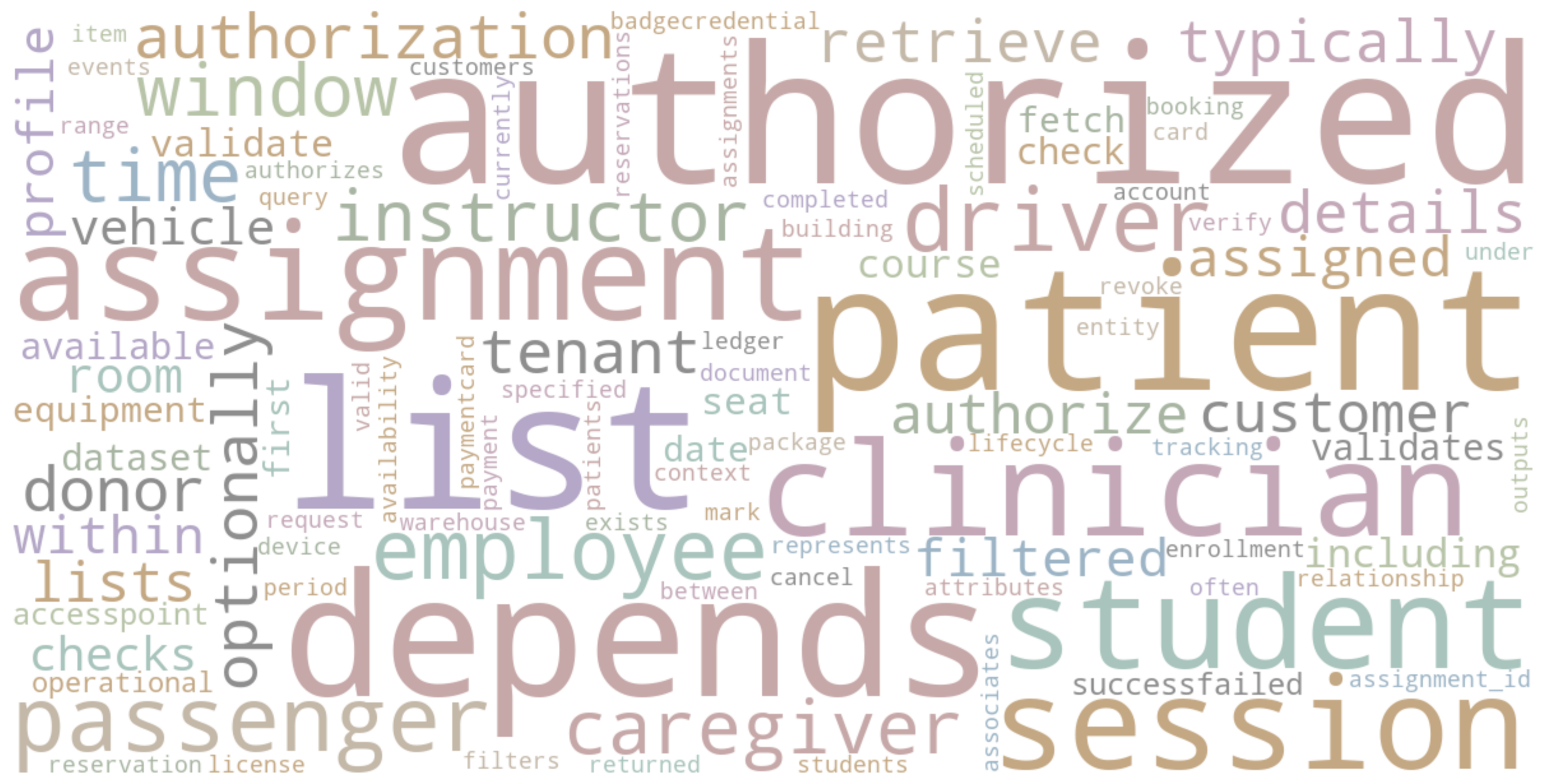} 
        \captionsetup{justification=centering, singlelinecheck=false}
        \caption{Semantic Keywords Cloud}
    \end{subfigure}
    \hfill
    \begin{subfigure}[b]{0.63\textwidth}
        \centering
        \includegraphics[width=\textwidth]{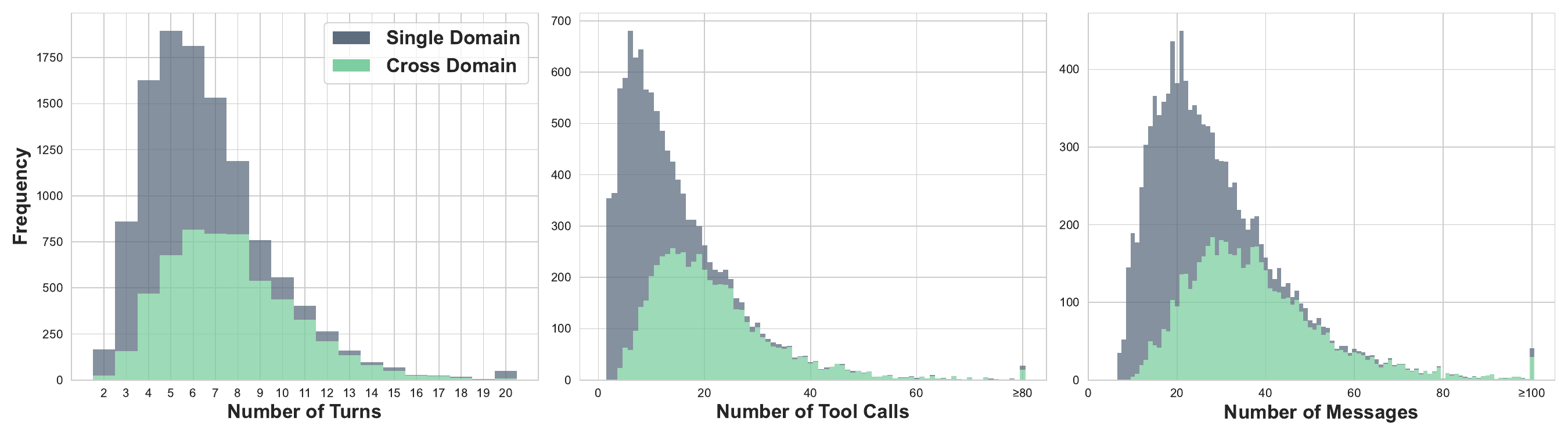} 
        \captionsetup{justification=centering, singlelinecheck=false}
        \caption{Complexity Metrics}
    \end{subfigure}
    \caption{\textbf{Complexity Analysis.} Cross-domain trajectories demonstrate significantly higher complexity in turn depth and tool usage compared to single-domain tasks.}
    \label{fig:complexity_stats}
\end{figure*}

A core objective of AgentSkiller is to simulate realistic, long-horizon, cross-domain interactions. The semantic analysis (Figure~\ref{fig:complexity_stats}(a)) reveals a prevalence of operational keywords such as ``authorized,'' ``assignment,'' and ``validate,'' confirming the tasks are grounded in rigorous policy execution. Furthermore, the complexity metrics (Figure~\ref{fig:complexity_stats}(b)) show that cross-domain tasks significantly extend the turn depth, with a long tail reaching 15--20 turns, requiring agents to orchestrate a larger set of functions to resolve composite user intents.

% --- Figure Group 2: Complexity ---

\subsection{Rigorous Quality Assurance and Error Taxonomy}
Unlike pipelines that prioritize yield over precision, AgentSkiller employs a strict \textit{Execution-Based Task Filtering} mechanism (Step 16). 

\begin{figure*}[t]
    \centering
    \begin{subfigure}[b]{0.6\textwidth}
        \centering
        \includegraphics[width=\textwidth]{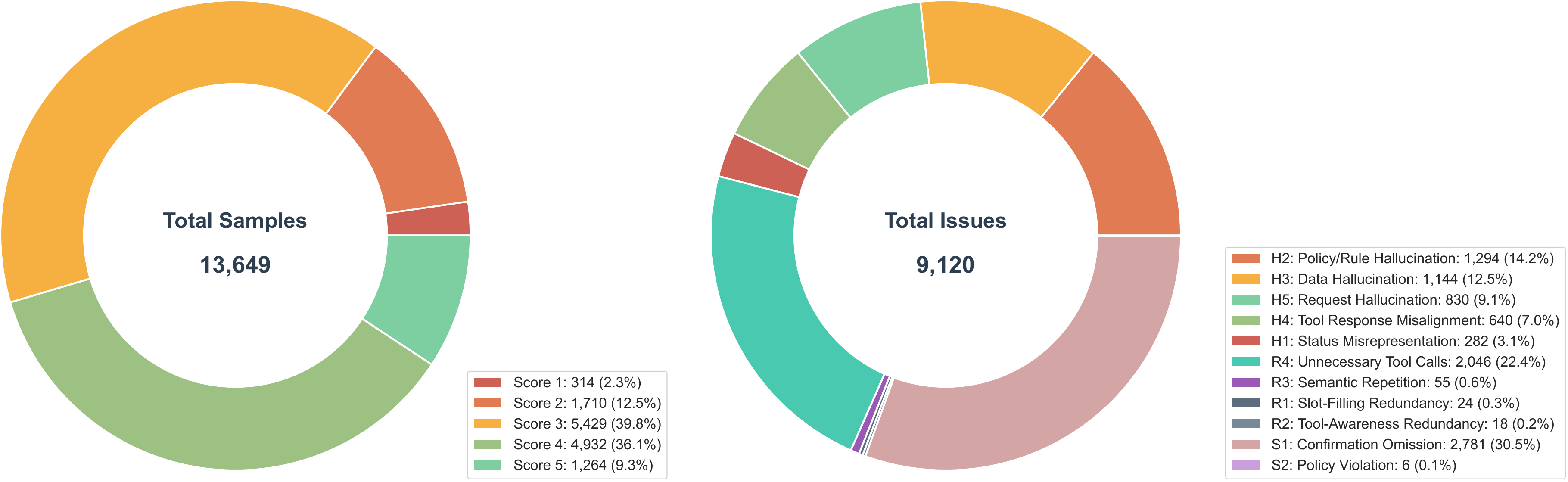}
        \captionsetup{justification=centering, singlelinecheck=false}
        \caption{Score Distribution \& Error Taxonomy}
        \label{fig:error_dist}
    \end{subfigure}
    \hfill
    \begin{subfigure}[b]{0.39\textwidth}
        \centering
        \includegraphics[width=\textwidth]{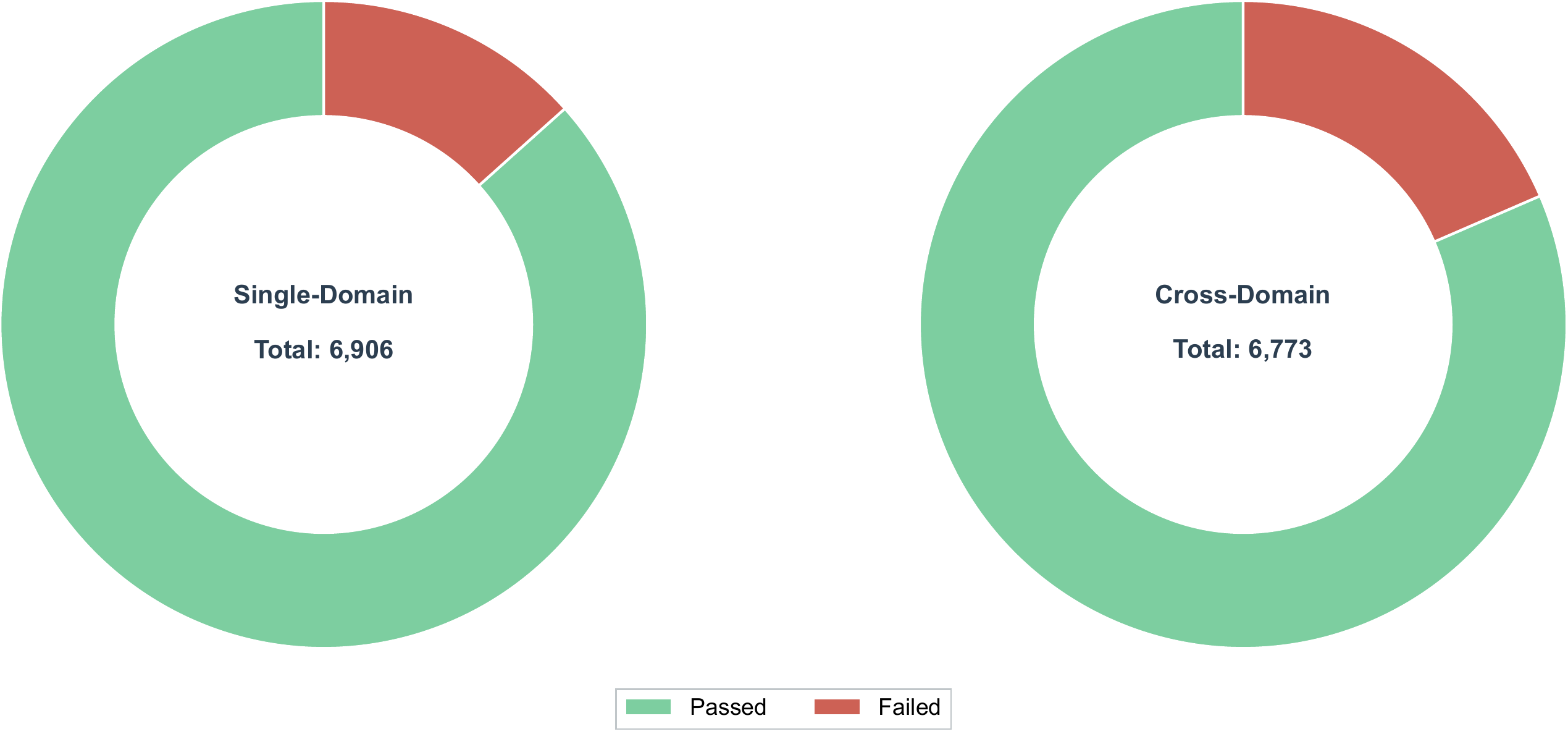}
        \captionsetup{justification=centering, singlelinecheck=false}
        \caption{Excluded Samples (Total: 1,887)}
        \label{fig:filtration}
    \end{subfigure}
    \caption{\textbf{Quality Assurance Statistics.} (a) The error taxonomy analyzes the issues found in the excluded samples, where the landscape is dominated by safety violations (S1) and efficiency bottlenecks (R4). (b) The donut charts of excluded samples demonstrate the rigorous, domain-agnostic filtration process across both domains.}
    \label{fig:quality_analysis}
\end{figure*}

\paragraph{Filtration Robustness.}
As shown in Figure~\ref{fig:quality_analysis}(b), \textbf{4,299 samples} (approximately 31.5\% of the total pool) were excluded for failing to meet the ``All-or-Nothing'' criteria. These exclusions are distributed across both single-domain and cross-domain tasks, serving as empirical evidence of the pipeline's unbiased rigor in eliminating execution failures.

\paragraph{Error Analysis.}
Figure~\ref{fig:quality_analysis}(a) breaks down the {9,120 specific issues} identified within these excluded samples. The distribution highlights the primary challenges for agents:
\begin{itemize}
    \item \textbf{Safety (S1):} \textit{Confirmation Omission} is the most prevalent error ($30.5\%$), reflecting the strict enforcement of safety protocols for state-mutating actions.
    \item \textbf{Efficiency (R4):} \textit{Unnecessary Tool Calls} ($22.4\%$) indicate that agents often struggle with optimal planning.
    \item \textbf{Hallucination (H2/H3):} Errors in \textit{Policy Hallucination} ($14.2\%$) and \textit{Data Hallucination} ($12.5\%$) confirm that the benchmark effectively tests context grounding.
\end{itemize}

% --- Figure Group 3: Quality & Filtration ---

\section{Experiments}

To assess the efficacy of our dataset, we apply supervised fine-tuning (SFT) on the models of varying sizes and analyzed their performance. We first detail the training setup, then proceed to evaluate these models relative to key baselines on popular function-calling benchmarks.

\subsection{Experimental Setup}
\begin{itemize}
    \item \textbf{Model and Fine-tuning Setup.} We perform SFT on Qwen3-4B, Qwen3-8B, and Qwen3-14B~\citep{yang2025qwen3}. The training is conducted using the verl~\citep{sheng2025hybridflow} framework with a sequence length of 128k, a learning rate of 1e-5, and a warmup ratio of 0.1. We utilize a cosine learning rate scheduler, setting the weight decay to 0.01, gradient clipping to 1.0, and optimizer betas to $(0.9, 0.95)$.
\item \textbf{Benchmarks.}
We evaluate our models on three widely used function-calling benchmarks:$\tau$-Bench~\citep{yao2024tau}, $\tau^{2}$-bench~\citep{barres2025tau} and ACEBench~\citep{chen2025acebench}. 
%==
For a comprehensive comparison, we incorporate results from both open-source models 
GPT-OSS-120B~\citep{agarwal2025gpt}, 
Deepseek-V3.1\footnote{\url{https://huggingface.co/deepseek-ai/DeepSeek-V3.1}}, 
Kimi-K2-1T~\citep{team2025kimi}, 
Qwen3-Thinking-235B~\citep{yang2025qwen3}, 
Seed-OSS-36B\footnote{\url{https://github.com/ByteDance-Seed/seed-oss}}, 
Qwen3-Coder-30B-A3B\footnote{\url{https://huggingface.co/Qwen/Qwen3-Coder-30B-A3B-Instruct}}, 
xLAM-2 series~\citep{prabhakar2025apigen}, and recent baselines including AgentScaler~\citep{fang2025towards}, 
ToolMind~\citep{yang2025toolmind}, 
EnvScaler~\citep{song2026envscaler}, 
MUA~\citep{zhao2025mua}, 
and ToolACE~\citep{liu2024toolace}; 
and closed-source models 
Gemini-2.5-Pro~\citep{comanici2025gemini}, 
Claude-Sonnet-4\footnote{\url{https://www.anthropic.com/news/claude-4}}, 
GPT-o3\footnote{\url{https://openai.com/zh-Hans-CN/index/introducing-o3-and-o4-mini/}}, 
GPT-o4-mini\footnote{\url{https://openai.com/zh-Hans-CN/index/gpt-4o-mini-advancing-cost-efficient-intelligence/}}, 
and GPT-5-think\footnote{\url{https://openai.com/zh-Hans-CN/index/introducing-gpt-5/}}.
%==
For $\tau$-Bench and $\tau^2$-Bench, we used GPT-5.2\footnote{\url{https://openai.com/zh-Hans-CN/index/introducing-gpt-5-2/}} as a user simulator and evaluate performance by directly
comparing the model before and after fine-tuning.
\item \textbf{Datasets.} The AgentSkiller-11K dataset includes 5,941 single-domain samples representing atomic tasks within isolated boundaries and 5,581 cross-domain samples capturing complex workflows across multiple services.
\end{itemize}

\subsection{Main Results}

% \definecolor{softblue}{RGB}{46, 116, 181}
\begin{table}[!t]
\centering
\caption{Model performance comparison on $\tau$-bench, $\tau^2$-bench and ACEBench-Agent. The best results are highlighted in \textbf{bold}, and the second-best results are \underline{underlined}.}
\resizebox{\textwidth}{!}{
\begin{tabular}{l|ccc|cccc|c}
\toprule
\multicolumn{1}{c|}{\textbf{Model}} & \multicolumn{3}{c|}{\textbf{$\tau$-Bench}} & \multicolumn{4}{c|}{\textbf{$\tau^2$-Bench}} & \textbf{ACEBench} \\
\cmidrule(lr){2-4} \cmidrule(lr){5-8} \cmidrule(lr){9-9}
& \textbf{Retail} & \textbf{Airline} & \textbf{Overall} & \textbf{Retail} & \textbf{Airline} & \textbf{Telecom} & \textbf{Overall} & \textbf{Agent} \\
\midrule

\multicolumn{9}{c}{\cellcolor{gray!20}\textbf{Closed-Source Large Language Models}} \\ 
\midrule
Gemini-2.5-pro & 68.7 & 44.0 & 61.2 & 67.5 & 56.0 & 27.2 & 48.9 & 63.4 \\
Claude-Sonnet-4 & \underline{73.9} & 40.0 & 63.6 & 67.5 & 54.0 & 47.4 & 56.8 & 42.5 \\
GPT-o3 & 70.4 & 52.0 & 64.8 & \underline{80.2} & \textbf{64.8} & 58.2 & 68.4 & 63.3 \\
GPT-o4-mini & 70.4 & 46.0 & 63.0 & 70.2 & 56.0 & 46.5 & 57.9 & 60.0 \\
GPT-5-think & \textbf{78.3} & 44.0 & \textbf{67.8} & \textbf{81.1} & \underline{62.6} & \textbf{96.7} & \textbf{84.2} & 32.5 \\
\midrule

\multicolumn{9}{c}{\cellcolor{gray!20}\textbf{Open-Source Large Language Models}} \\
\midrule
GPT-OSS-120B-A5B & 67.8 & 49.2 & 62.1 & 57.0 & 38.0 & 45.6 & 48.9 & 50.8 \\
Deepseek-V3.1-671B-A37B & 66.1 & 40.0 & 58.1 & 64.9 & 46.0 & 38.5 & 50.7 & 40.8 \\
Kimi-K2-1T-A32B & \underline{73.9} & 51.2 & 67.0 & 70.6 & 56.5 & 65.8 & 66.1 & 65.0 \\
Qwen3-Thinking-235B-A22B & 67.8 & 46.0 & 61.2 & 71.9 & 58.0 & 45.6 & 58.6 & 39.1 \\
Seed-OSS-36B & 70.4 & 46.0 & 63.0 & 68.4 & 52.0 & 41.2 & 54.3 & 58.4 \\
Qwen3-Coder-30B-A3B & 68.7 & 48.0 & 62.4 & 60.5 & 42.0 & 30.7 & 45.0 & 24.1 \\
xLAM-2-8B-fc-r & 58.2 & 35.2 & 51.2 & 55.3 & 48.0 & 11.4 & 36.0 & 5.0 \\
xLAM-2-32B-fc-r & 64.3 & 45.0 & 58.4 & 55.3 & 52.0 & 16.7 & 38.9 & 13.4 \\
xLAM-2-70B-fc-r & 67.1 & 45.2 & 60.4 & 61.4 & 56.0 & 14.0 & 41.0 & 38.4 \\
Qwen3-Thinking-4B & 59.1 & \underline{52.5} & 57.1 & 56.1 & 52.0 & 28.7 & 44.1 & 11.7 \\
Qwen3-8B & 45.2 & 25.0 & 39.0 & 41.2 & 30.5 & 23.5 & 32.0 & 29.1 \\
Qwen3-14B & 45.7 & 31.0 & 41.2 & 48.0 & 30.0 & 26.9 & 36.1 & 44.2 \\
Qwen3-Thinking-30B-A3B & 67.8 & 48.0 & 61.8 & 58.8 & 58.0 & 26.3 & 45.3 & 42.8 \\
AgentScaler-4B & 64.3 & \textbf{54.0} & 61.2 & 62.3 & 56.0 & 48.2 & 55.4 & 30.8 \\
AgentScaler-8B & 50.4 & 42.0 & 47.8 & 58.8 & 44.0 & 45.4 & 50.6 & 44.2 \\
AgentScaler-30B-A3B & 70.4 & \textbf{54.0} & 65.4 & 70.2 & 60.0 & 55.3 & 62.3 & 60.0 \\
ToolMind-8B & 57.4 & 36.0 & 50.9 & 59.7 & 48.0 & 31.6 & 46.1 & - \\
ToolMind-14B & 60.0 & 46.0 & 55.7 & 59.7 & 56.0 & 31.6 & 47.5 & - \\
EnvScaler-4B (w/ SFT) & 44.4 & 40.0 & 43.1 & 46.5 & 36.0 & 23.7 & 35.3 & 40.0 \\
EnvScaler-8B (w/ SFT) & 48.7 & 34.0 & 44.2 & 43.0 & 40.0 & 26.3 & 35.6 & 70.0 \\
EnvScaler-14B (w/ SFT) & 55.7 & 26.0 & 46.6 & 49.1 & 42.0 & 35.7 & 42.3 & 64.0 \\
MUA-4B (w/ SFT) & 14.8 & 18.0 & 15.8 & 11.4 & 36.0 & 21.2 & 19.8 & 40.0 \\
MUA-8B (w/ SFT) & 43.5 & 26.0 & 38.2 & 47.4 & 40.0 & 23.9 & 36.4 & 66.0 \\
MUA-14B (w/ SFT) & 48.7 & 36.0 & 44.8 & 57.0 & 42.0 & 55.3 & 53.6 & 70.0 \\
ToolACE-4B & 13.0 & 14.0 & 13.3 & 12.3 & 24.0 & 20.2 & 17.6 & 26.0 \\
ToolACE-8B & 27.0 & 10.0 & 21.8 & 38.7 & 18.0 & 21.2 & 27.8 & 50.0 \\
ToolACE-14B & 37.4 & 22.0 & 32.7 & 36.8 & 24.0 & 29.8 & 31.6 & 52.0 \\
\midrule

\multicolumn{9}{c}{\cellcolor{gray!20}\textbf{AgentSkiller (Single-Domain Subset)}} \\
\midrule
AgentSkiller-SD-4B & 47.0 & 24.0 & 40.0 & 52.6 & 34.0 & 72.6 & 57.5 & 64.0 \\
AgentSkiller-SD-8B & 53.9 & 32.0 & 47.2 & 51.3 & 42.0 & 61.1 & 53.6 & \underline{76.0} \\
AgentSkiller-SD-14B & 53.0 & 28.0 & 45.4 & 69.3 & 44.0 & 85.1 & 71.2 & \textbf{78.0} \\
\midrule

\multicolumn{9}{c}{\cellcolor{gray!20}\textbf{AgentSkiller (Full Set (with Cross-Domain Subset))}} \\
\midrule
AgentSkiller-4B & 62.6 & \textbf{54.0} & 60.0 & 60.6 & 54.0 & 76.6 & 66.0 & 54.0 \\
AgentSkiller-8B & 59.1 & 36.0 & 52.1 & 59.7 & 46.0 & 67.0 & 60.2 & 68.0 \\
\textbf{AgentSkiller-14B} & {73.0} & \textbf{54.0} & \underline{{67.2}} & {77.2} & \textbf{56.0} & \underline{{91.2}} & \underline{{79.1}} & \textbf{78.0} \\
\bottomrule
\end{tabular}
}
\end{table}

\noindent
\textbf{Competitive Performance Against Proprietary Models.}
The experimental results presented in Table 1 demonstrate that AgentSkiller-14B establishes a new standard for open-source tool-learning models, achieving performance parity with, and in some cases surpassing, top-tier closed-source systems. On the complex $\tau^2$-bench, which evaluates multi-turn and cross-domain capabilities, AgentSkiller-14B achieves an overall score of 79.1\%, significantly outperforming GPT-o3 with 68.4\% and Claude-Sonnet-4 at 56.8\%. Notably, in the Telecom domain, the model attains a score of 91.2\%, securing the second-best position globally and trailing only the much larger GPT-5-think. Furthermore, on ACEBench-Agent, AgentSkiller-14B records a score of 78.0\%, surpassing all compared open-source baselines and outperforming Gemini-2.5-pro by a margin of 14.6\%. This indicates that high-quality synthetic data can effectively bridge the capability gap between compact open-source models and proprietary giants.

\noindent
\textbf{Efficacy of Cross-Domain Data Synthesis.}
A comparative analysis between the Single-Domain (SD) subset and the Full Data configuration reveals the critical importance of diverse training trajectories. While the AgentSkiller-SD-14B model performs strongly on ACEBench-Agent with a score of 78.0\%, its performance on the more intricate $\tau^2$-bench is limited to 71.2\%. The integration of cross-domain data in the Full configuration elevates this score to 79.1\%, with the most substantial gains observed in the Retail domain, improving from 69.3\% to 77.2\%. More strikingly, on the $\tau$-bench, the transition from SD to Full Set yields a performance leap from 45.4\% to 67.2\%. This trend suggests that the cross-domain synthesis pipeline not only enhances the model's ability to handle complex, interleaved tasks but also reinforces its robustness on fundamental tool-use scenarios by exposing the model to a broader distribution of domain environment interaction patterns.

\noindent
\textbf{Parameter Efficiency and Scaling Dynamics}
The data highlights a compelling trade-off between model scale and data quality. AgentSkiller-4B, despite its compact size, achieves an overall $\tau^2$-bench score of 66.0\%, remarkably outperforming the significantly larger xLAM-2-70B-fc-r at 41.0\% and Qwen3-Thinking-235B scoring 58.6\%. This result challenges the prevailing assumption that complex reasoning requires massive parameter counts, instead pointing to data quality as the governing factor in function-calling proficiency. Additionally, we observe consistent scaling laws within the AgentSkiller family on complex benchmarks. On the Telecom task, performance scales monotonically from 76.6\% for the 4B model to 91.2\% for the 14B variant, confirming that the proposed data synthesis method provides a stable learning signal that benefits from increased model capacity.

\subsection{Analysis Results}
\begin{figure}[!t] % h-此处,t-顶部,b-底部,p-单独一页
    \centering
        \centering      
        \includegraphics[width=\linewidth]{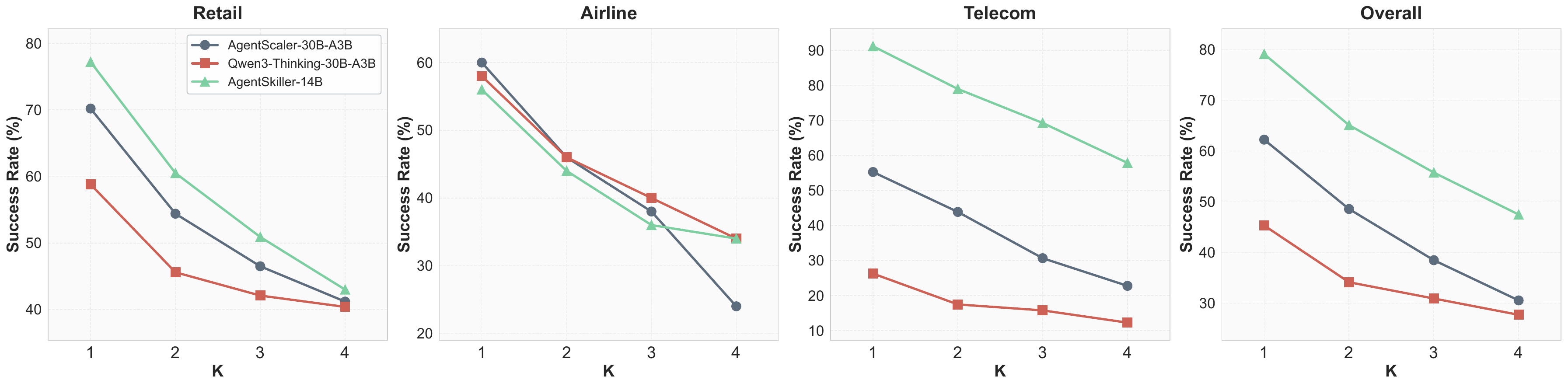}
        % \captionsetup{justification=centering, singlelinecheck=false}
        \caption{Pass\^{}k metric results across all domains in the $\tau^2$-Bench of our AgentSkiller-14B compared with AgentScaler-30B-A3B and Qwen3-Thinking-30B-A3B.}
        \label{fig:passk}
\end{figure}
\textbf{Robustness and Consistency Analysis.}
To further evaluate the stability of model performance under varying success criteria, we analyze the metric on the $\tau^2$-bench, as illustrated in Figure~\ref{fig:passk}. This evaluation pits our AgentSkiller-14B against significantly larger baselines, specifically AgentScaler-30B-A3B and Qwen3-Thinking-30B-A3B. The results reveals a decisive advantage for AgentSkiller-14B across the majority of domains, despite it possessing less than half the parameter count of the competing models. In the Overall and Retail categories, AgentSkiller-14B maintains a distinct performance margin across all  values, demonstrating superior consistency. The contrast is most pronounced in the Telecom domain, where AgentSkiller-14B initiates with a success rate near 90\% at  and maintains a robust lead as  increases, whereas the 30B parameter baselines exhibit a sharp performance degradation, starting below 60\%. While the performance in the Airline domain remains tightly clustered among all models, the aggregate trends confirm that the high-quality synthetic data generated by our framework imparts a level of execution robustness that outweighs raw model scale.

% \begin{figure}[!t] % h-此处,t-顶部,b-底部,p-单独一页
%     \centering
%         \centering      
%         \includegraphics[width=\linewidth]{figures/tool_calls_domain_comparison.pdf}
%         \captionsetup{justification=centering, singlelinecheck=false}
%         \caption{Accuracy by tool call count of AgentSkiller-14B across different domains on $\tau^2$-bench.}
%         \label{fig:tool_call_domain}
% \end{figure}

\begin{figure}[!t]
    \centering
    \includegraphics[width=\linewidth]{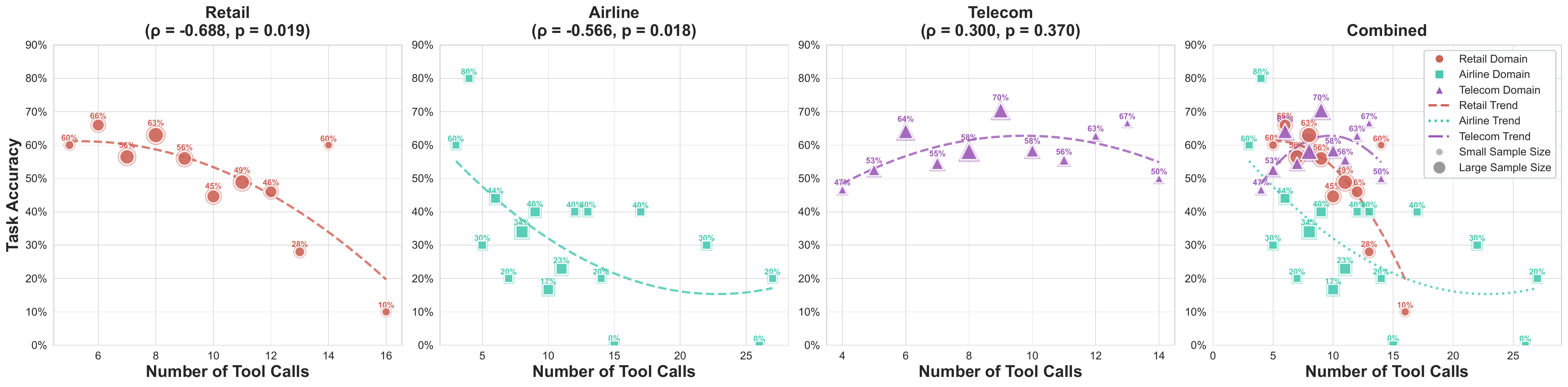} % 替换为你的第二张PDF
    \captionsetup{justification=centering, singlelinecheck=false}
    \caption{Accuracy by tool call count of AgentSkiller-14B across different domains on $\tau^2$-bench.}
    \label{fig:tool_call_domain}
\end{figure}

\noindent
\textbf{Performance Scaling with Interaction Depth.}
Figure~\ref{fig:tool_call_domain} depicts the fine-grained relationship between task execution accuracy and interaction complexity, measured by the number of tool calls, across the three domains of the $\tau^2$-bench. A distinct divergence in performance trajectories is observed. For the Retail and Airline domains, the model exhibits a statistically significant negative correlation between trajectory length and success rate, with Pearson correlation coefficients $\rho$ of -0.688 at $p=0.019$ and -0.566 at $p=0.018$, respectively. Specifically, in the Retail domain, accuracy remains robust above 60\% for short-horizon tasks requiring 6-8 tool calls but experiences a precipitous decline to below 20\% as the sequence length extends beyond 14 steps. Similarly, the Airline domain shows high initial proficiency reaching 80\% at low complexity but suffers from error accumulation in long-horizon scenarios, with accuracy dropping to near zero for tasks requiring more than 20 tool calls.

\noindent
\textbf{Domain-Specific Robustness Profiles.}
In sharp contrast to the degradation observed in Retail and Airline tasks, the Telecom domain displays an anomalous and resilient performance profile, characterized by a positive correlation coefficient of $\rho = 0.300$. Rather than decaying with increased complexity, the model's accuracy initially ascends, peaking at 70\% around 8 tool calls, and maintains a competitive performance level approximately 50-60\% even as the interaction depth extends to 12-14 calls. This inverted U-shaped trend suggests that the AgentSkiller-14B model has effectively internalized the complex, multi-step procedural logic inherent to Telecom workflows such as diagnostic troubleshooting or plan modification, where longer tool sequences often represent necessary, deterministic steps rather than improved confusion or error recovery. The combined visualization further highlights this stratification, showing that while Airline tasks present the highest difficulty ceiling at extreme lengths, the model demonstrates exceptional stability in the Telecom domain, resisting the typical trend of long-context performance collapse.

\section{Conclusion}
In this work, we introduced \textbf{AgentSkiller}, a scalable framework for synthesizing high-fidelity, cross-domain executable agent environments. By rigorously enforcing state-machine determinism and logical consistency, AgentSkiller automates the generation of diverse tool-use trajectories that mirror real-world complexity. 
%==
A pivotal finding of our study is the indispensable role of \textbf{semantically integrated cross-domain synthesis}. Our experiments on the $\tau^2$-bench reveal that single-domain training alone is insufficient for generalist intelligence; it is the fusion of distinct service blueprints into interleaved, long-horizon workflows that effectively unlocks an agent's ability to handle complex planning and state management. The empirical success of AgentSkiller-14B, which outperforms significantly larger proprietary models, validates a critical paradigm shift: scaling generalist agent capabilities depends less on sheer parameter count and more on the structural depth and logical connectivity of the training data. We hope AgentSkiller serves as a foundational platform for future research in robust, autonomous generalist agents.

\bibliography{main}

\newpage
\appendix
\section*{Appendix}
\section{Tool Graph Visualization}
\begin{figure}[H]
    \centering
    \includegraphics[width=0.76\linewidth]{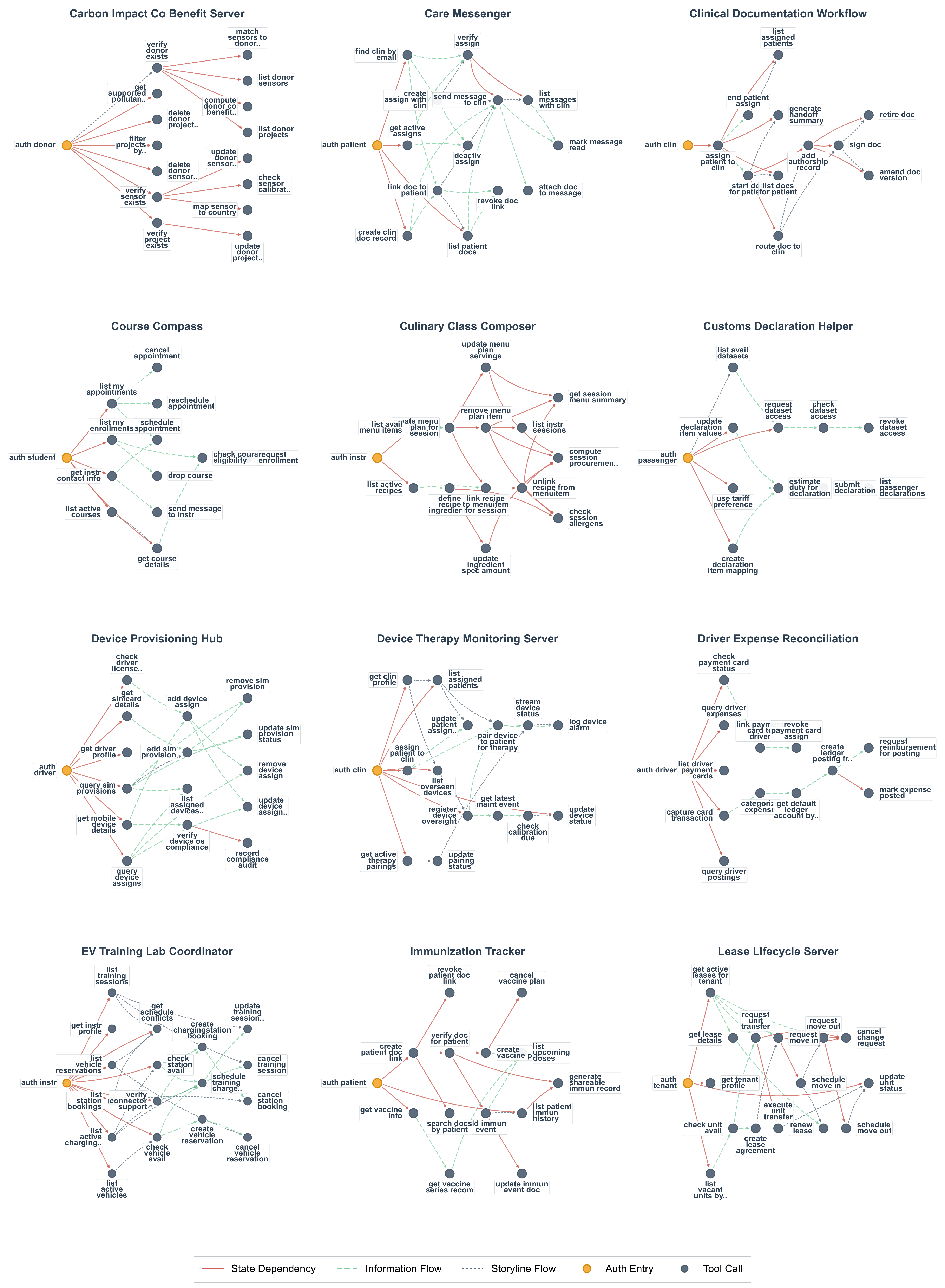}
     \captionsetup{justification=centering, singlelinecheck=false}
    \caption{Case studies of Tool Graph visualization across different domains.}
    \label{fig:tool_graph_cases}
\end{figure}

Figure~\ref{fig:tool_graph_cases} visualizes the structural diversity of generated Tool Graphs. Each graph is anchored by an orange root \textit{Auth Entry} node establishing user identity and initial state. The topology expands into a DAG of gray \textit{Tool Call} nodes governed by three edge types. Red solid lines denote state dependencies enforcing execution prerequisites, such as requiring \texttt{create\_clin\_doc\_record} to precede \texttt{link\_doc\_to\_patient} in \textit{Care Messenger}. Green dashed lines represent information flow, handling dynamic parameter passing like \texttt{patient\_id} across actions. Finally, black dotted lines indicate storyline flow, modeling probabilistic transitions to introduce realistic noise and non-deterministic branching. The structural variation, ranging from shallow, broad trees in \textit{Carbon Impact Co-Benefit} to deep, intertwined chains in \textit{Lease Lifecycle Server}, confirms the framework's ability to autonomously construct complex, domain-specific execution logic.

\section{Core Prompts}
% === 颜色定义 ===
\definecolor{lowsatblue}{RGB}{94, 129, 172} 
\definecolor{bglight}{RGB}{252, 252, 252}

% === Box 样式定义 (带 Caption 版) ===
% [auto counter] 开启自动编号
% [1] 表示接受 1 个参数作为标题
\newtcblisting[auto counter]{promptbox}[1]{
    enhanced,
    breakable,
    listing only,
    % === 标题设置 ===
    title={#1}, % 显示 "Box 1: 你的标题"
    coltitle=white,           % 标题文字颜色
    colbacktitle=lowsatblue,  % 标题背景颜色 (使用你的蓝色)
    fonttitle=\bfseries,      % 标题字体加粗
    attach boxed title to top left={xshift=0mm, yshift=0mm}, % 标题位置修正
    boxed title style={sharp corners, size=small, frame hidden}, % 标题框样式
    % === 盒子主体设置 ===
    colback=bglight,
    colframe=lowsatblue,      % 边框颜色与标题一致，整体感更强
    boxrule=0.5pt,
    arc=2pt,                  %稍微圆角一点点
    left=5mm, right=5mm, top=3mm, bottom=3mm,
    % === 代码设置 (保持全黑 + 手动高亮) ===
    listing options={
        basicstyle=\ttfamily\footnotesize,
        breaklines=true,
        columns=fullflexible,
        keepspaces=true,
        numbers=none,
        escapechar=^,
        showstringspaces=false,
        language={},
        keywordstyle=\color{black},
        stringstyle=\color{black},
        commentstyle=\color{black},
        identifierstyle=\color{black}
    }
}

% === 变色辅助命令 ===
\newcommand{\ph}[1]{\textcolor{lowsatblue}{#1}}
% === 这里是正文中的使用 ===
\begin{promptbox}{Blueprint Generation Prompt}
You are implementing a detailed MCP server blueprint based on the given outline.

<outline>
Server Name: ^\ph{\{server\_name\}}^
Description: ^\ph{\{description\}}^
Entities: ^\ph{\{entities\}}^
</outline>

<entity_definitions>
^\ph{\{entity\_definitions\}}^
</entity_definitions>

---

### Step by Step Generation
1. The core entity (first in the list) is the user interacting with the LLM agent.
2. Design relationships between the entities.
   - All entities and relationships are physically implemented as dataframes.
   - Relationships' attributes are columns of inter-entity interaction dataframe.
   - Relationship should have its own ID, NOT reference other relationships' IDs.
   - The core entity must be involved in all relationships. In other words, relationship that ONLY involves peripheral entities and not involve the core entity is NOT allowed.
   - Relationship attributes should ONLY reference entity attributes as foreign keys, NOT other relationships' keys. Each relationship connects entities directly (e.g., a relationship can have student_id and course_id, but should NOT have enrollment_id pointing to another relationship).
3. Design functions that operate the relationship dataframe (query records, add record, update record, etc).
4. Design auxiliary functions that complete the MCP server functionality.
   - For example, `list_all_courses()` in a `CourseManagement` MCP server.

### Requirements
- Only up to ^\ph{\{max\_relationships\}}^ relationships are allowed between two entities.
- The MCP server should include at least ^\ph{\{min\_functions\}}^ functions.
- Function parameters should NOT assume relationships or entities that do not exist.
- An authorization function (only one) is needed to authorize the user (core entity) with its ID at the beginning. The authorization function should ONLY return Success/Failed status, NOT return any tokens (e.g., auth_token). The session context will manage authentication state internally.
- You are NOT allowed to add new attributes to entities.
- You are NOT allowed to add new entities beyond those specified.
- There should be dependencies among functions to increase interaction complexity:
  - Input parameters of core functions should rely on outputs of other relevant functions.
  - For example, before enrolling a Student to a Course, check prerequisites first.

### Output Format
Return your answer in JSON format:
```json
{{
  "MCP_server_name": "^\ph{\{server\_name\}}^",
  "description": "^\ph{\{description\}}^",
  "core_entity": "^\ph{\{core\_entity\}}^",
  "peripheral_entities": ^\ph{\{peripheral\_entities\_json\}}^,
  "relationships": [
    {{
      "name": "",
      "description": "",
      "attributes": {{
        "attr_name": {{
          "type": "",
          "value_from_entity": "",
          "range": ""
        }}
      }}
    }}
  ],
  "functions": [
    {{
      "name": "",
      "description": "",
      "legal_accessor": [],
      "parameters": {{
        "param_name": {{
          "description": "",
          "type": "",
          "range": ""
        }}
      }}
    }}
  ]
}}
Output only the JSON object, no additional text.
\end{promptbox}

\begin{promptbox}{Tool-Graph Generation Prompt}
You are an expert in reasoning about tool dependencies in modular agent systems.
Given a set of tool descriptions for ^\ph{\{domain\_name\}}^,

<tools>
^\ph{\{tools\}}^
</tools>

and the policy of the system

<policy>
^\ph{\{policy\}}^
</policy>

your task is to construct a **Tool Execution Graph**.

### Guidelines
1. Each node represents one tool function.
2. Each edge represents an **execution dependency**.
3. Use directed edges (`"source" -> "target"`) for:
    - **State Dependencies** (e.g., authentication, record retrieval)
    - **Information Flow** (output of one tool is required by another)
    - **Storyline Flow** (a tool is usually followed by another tool in real-world scenario).
4. Never omit the nodes list -- ensure every mentioned tool exists as a node.
5. **Single Source of Truth:** The node `Authorize` must be the **ONLY** node with an In-Degree of 0. It is the starting point of the entire workflow.
6. **Full Connectivity:** Every node (except `Authorize`) must have at least one incoming edge (In-Degree >= 1). This ensures all nodes are reachable from `Authorize`.
7. **Acyclicity:** The graph must strictly be a DAG. No circular dependencies (e.g., A->B->A is forbidden).
8. **Storyline Flow:** Consider this connection ONLY when the full connectivity constraint CANNOT be satisfied by the other two types of dependencies.
8. The output must be valid JSON that can be loaded by networkx.
9. Output JSON in the following schema without any extra explanations.
```json
{{
    "directed": true,
    "multigraph": false,
    "nodes": [
        {{"id": "tool_name"}},
        ...
    ],
    "links": [
        {{"source": "tool_a", "target": "tool_b", "type": "state dependency"}},
        {{"source": "tool_a", "target": "tool_c", "type": "information flow"}},
        {{"source": "tool_a", "target": "tool_d", "type": "storyline flow"}},
        ...
    ]
}}
```
\end{promptbox}

\begin{promptbox}{MCP Server Implementation Instruction}
You must implement an MCP server as a complete Python module.

Produce the complete MCP server implementation according to the domain policy,
database blueprint, MCP server code template, and the behavioral constraints described as follows.

=====================
## 0. Inputs
=====================

### 0.1 Domain Policy
^\ph{\{domain\_policy\}}^

### 0.2 MCP Server Blueprint
^\ph{\{blueprint\}}^

### 0.3 MCP Server Code Template
^\ph{\{mcp\_server\_code\_template\}}^

=====================
## 1. Implementation Target
=====================

Generate a Python file containing:
- A session context @dataclass
  - Session state management
  - Session database isolation (independent database per session_id)
- A full MCP server class
  - All tools implemented as class methods
  - A unified router entrypoint: invoke(session_id, tool_name, **kwargs)
    - 'session_id' helps you find the session context and database copies
    - 'tool_name' is the tool you would like to call
    - '**kwargs' are the parameters of the tool you would like to call
  - Domain-policy violation checking
- All "xx_id" must satisfy ^\ph{\{\{entity\_name.lower()\}\}}^\_id.

### 1.1 Interface
The implemented MCP server must be compatible with the standard wrapper interface.

### 1.2 Load Database
Get database root by
'''python
script_dir = os.path.dirname(os.path.abspath(__file__))
db_root = os.path.abspath(os.path.join(script_dir, '..', 'database', 'outputs'))
'''

**NOTE**
- '__init__(self, domain_name)' should call '_load_database(domain_name)' during initialization.
- Never mock new databases instead of reading from the given files.
- The current time is ^\ph{\{simulation\_time\}}^.

### 1.3 Output Rules
- All tools must return JSON-serializable Python objects
- No printing
- No placeholder methods

=====================
## 2. CRITICAL IMPLEMENTATION RULES
=====================

### 2.1 Confirmation Pattern (REQUIRED for data-changing functions)
If the policy mentions that a function requires "user confirmation" before execution:
- Add a 'confirm: bool = False' parameter to the function
- When 'confirm=False': Return a preview of what will happen with '{{"needs_confirmation": True, "action_preview": "...description..."}}'
- When 'confirm=True': Execute the actual action and return the result
- The LLM Assistant will first call with confirm=False, show the preview to user, then call again with confirm=True after user says "yes"

### 2.2 Accept Extra Parameters (REQUIRED for ALL functions)
Every function MUST accept '**kwargs' as the last parameter to gracefully ignore unexpected parameters.

### 2.3 Core Entity ID Handling
- Authorization binds the core entity ID to the session
- After authorization, functions should use 'session.authorized_^\ph{\{\{core\_entity\}\}}^_id' instead of requiring it as a parameter

=====================
## 3. Output Constraints
=====================

1. Do NOT include unit tests.
2. The final output must be valid Python code (no markdown).
3. All imports must be included.
4. The module must execute without modification.
5. EVERY function must end with '**kwargs' in its signature.
6. Data-changing functions MUST implement the confirm pattern if policy requires confirmation.

Output only the Python code, no markdown formatting.
\end{promptbox}

\begin{promptbox}{MCP Server Error Tracing Prompt}
I'm building an MCP server following the policy below
<policy>
^\ph{\{policy\}}^
</policy>

Here is the current implementation:
'''python
^\ph{\{mcp\_server\_code\}}^
'''

Such code failed in the following unit test:
<unit_test_failure>
^\ph{\{unit\_test\_failure\}}^
</unit_test_failure>

The test data generator:
<test_data_generator>
^\ph{\{test\_data\_generator\}}^
</test_data_generator>

## Task
Identify the reason that caused the failure:
- Is it because of wrong implementation of the MCP server?
- Is it simply a wrong unit test?
- Is it related to improper test data selection?

IMPORTANT: You MUST respond with ONLY a valid JSON object. No additional text before or after.

'''json
{{
    "failed_test_name": "<exact test function name>",
    "likely_bug_location": "<one of: function | unit test | test data>",
    "explanation": "<brief explanation of what's wrong>"
}}
'''

Respond with ONLY the JSON object above, nothing else.
\end{promptbox}

\setcounter{figure}{0}
\makeatletter 
\renewcommand{\thefigure}{A\@arabic\c@figure}
\makeatother

\setcounter{table}{0}
\makeatletter 
\renewcommand{\thetable}{A\@arabic\c@table}
\makeatother

\end{document}